\documentclass[journal]{IEEEtran}

\ifCLASSINFOpdf
   \usepackage[pdftex]{graphicx}
\else
   \usepackage[dvips]{graphicx}
\fi
\newcommand{\figsuffix}{png}

\usepackage{amsmath}
\usepackage{mathrsfs}
\usepackage{stmaryrd}

\usepackage{booktabs}
\usepackage{multirow}
\usepackage{arydshln}

\makeatletter
\def\adl@drawiv#1#2#3{%
        \hskip.5\tabcolsep
        \xleaders#3{#2.5\@tempdimb #1{1}#2.5\@tempdimb}%
                #2\z@ plus1fil minus1fil\relax
        \hskip.5\tabcolsep}
\newcommand{\cdashlinelr}[1]{%
  \noalign{\vskip\aboverulesep
           \global\let\@dashdrawstore\adl@draw
           \global\let\adl@draw\adl@drawiv}
  \cdashline{#1}
  \noalign{\global\let\adl@draw\@dashdrawstore
           \vskip\belowrulesep}}
\makeatother

\usepackage{url}

\usepackage{hyperref}
\hypersetup{
    colorlinks=true, % for debugging.
    linkcolor=blue,
    filecolor=magenta,      
    urlcolor=cyan,
}

\usepackage[capitalise]{cleveref}

\usepackage{bm,bbm}
\usepackage{amsfonts}
\usepackage{scrextend}
\usepackage{cancel}
\usepackage{epsfig}
\usepackage{amssymb}

\usepackage{soul}
\usepackage{listings}
\usepackage{caption}
\usepackage{subcaption}

\newcommand\ssquare[0]{\mathbin{\vcenter{\hbox{\scalebox{0.75}{$\blacksquare$}}}}}
\newcommand\sdiamond[0]{\mathbin{\vcenter{\hbox{\rotatebox{45}{\scalebox{0.70}{$\blacksquare$}}}}}}

\usepackage{color}
\usepackage{xcolor}
\usepackage{colortbl}

\definecolor{amber}{HTML}{E15D02}
\definecolor{amazonite}{HTML}{339C84}
\definecolor{ruby}{HTML}{E02020}
\definecolor{emerald}{HTML}{6DD400}
\definecolor{lgreen}{HTML}{5fcf5f} %7ED47E
\definecolor{lred}{HTML}{db6063} %E37E7F
\definecolor{citrine}{HTML}{F7B500}
\definecolor{ioite}{RGB}{98, 54, 255}
\definecolor{ruby}{RGB}{224, 32, 32}
\definecolor{greycol}{RGB}{230, 230, 230}
\definecolor{ametyst}{RGB}{182, 32, 224}
\definecolor{pink}{RGB}{213, 0, 115}
\definecolor{amber}{RGB}{250, 100, 0}
\definecolor{amazonite}{RGB}{68, 215, 182}
\definecolor{graphite}{RGB}{109, 114,120}
\definecolor{lightgrey}{RGB}{250, 250, 250}

\newcommand\paretocolor{lgreen!20}

\usepackage{amsthm}
\theoremstyle{definition}
\newtheorem*{definition}{Definition}

\usepackage[numbers,sort]{natbib}

\begin{document}

% \title{Continual Spatio-Temporal Graph Convolutional Neural Networks \\ for Online Activity Recognition}
% \title{Continual Skeletons for Efficient Online Activity Recognition}
% \title{Continual Skeletons: Reducing the Complexity of Online Skeleton-based Action Recognition}
% \title{Continual Spatio-Temporal GCN: Reducing the Cost of Online Skeleton-based Action Recognition}
% \title{Reducing the Complexity of Online Skeleton-based Action Recognition through Continual Inference}
% \title{Reducing the Complexity of Online Skeleton-based Action Recognition with Continual Inference Networks}
% \title{Reducing the Complexity of Online Skeleton-based Action Recognition with Continual ST-GCNs}
% \title{Online Skeleton-based Action Recognition \\with Continual Spatio-Temporal Graph Convolutional Networks}
\title{Continual Spatio-Temporal\\Graph Convolutional Networks}
% \title{Continual Skeletons: \\ Super Real-Time Online Skeleton-based Action Recognition}
% \title{Reducing the Complexity of Graph Convolutional Networks for Online Skeleton-based Action Recognition}


\author{
    \IEEEauthorblockN{
        Lukas Hedegaard,
        Negar Heidari, and 
        Alexandros Iosifidis
    }\\
    \IEEEauthorblockA{
        Department of Electrical and Computer Engineering, Aarhus University, Denmark\\
        \small{
        \texttt{\{lhm, negar.heidari, ai\}@ece.au.dk}
        }
    }
}


\maketitle

% \begingroup\renewcommand\thefootnote{*}
% \footnotetext{Equal contribution}
% \endgroup

% As a general rule, do not put math, special symbols or citations in the abstract

\begin{abstract}
Graph-based reasoning over skeleton data has emerged as a promising approach for human action recognition.
However, the application of prior graph-based methods, which predominantly employ whole temporal sequences as their input, to the setting of online inference entails considerable computational redundancy.
In this paper, we tackle this issue by reformulating the Spatio-Temporal Graph Convolutional Neural Network as a Continual Inference Network, which can perform step-by-step predictions in time without repeat frame processing.
To evaluate our method, we create a continual version of ST-GCN, \textit{Co}ST-GCN, alongside two derived methods with different self-attention mechanisms, \textit{Co}AGCN and \textit{Co}S-TR. We investigate weight transfer strategies and architectural modifications for inference acceleration, and perform experiments on the NTU RGB+D 60, NTU RGB+D 120, and Kinetics Skeleton 400 datasets. Retaining similar predictive accuracy, we observe up to 109$\times$ reduction in time complexity, on-hardware accelerations of 26$\times$, and reductions in maximum allocated memory of 52\% during online inference.
\end{abstract}

% Graph-based reasoning over skeleton data has emerged as a promising approach for human action recognition. However, the application of prior graph-based methods, which predominantly employ whole temporal sequences as their input, to the setting of online inference entails considerable computational redundancy. In this paper, we tackle this issue by reformulating the Spatio-Temporal Graph Convolutional Neural Network as a Continual Inference Network, which can perform step-by-step predictions in time without repeat frame processing. To evaluate our method, we create a continual version of ST-GCN, CoST-GCN, alongside two derived methods with different self-attention mechanisms, CoAGCN and CoS-TR. We investigate weight transfer strategies and architectural modifications for inference acceleration, and perform experiments on the NTU RGB+D 60, NTU RGB+D 120, and Kinetics Skeleton 400 datasets. Retaining similar predictive accuracy, we observe up to 109x reduction in time complexity, on-hardware accelerations of 26x, and reductions in maximum allocated memory of 52% during online inference.

\begin{IEEEkeywords}
Graph Convolutional Networks, Continual Inference, Efficient Deep Learning, Skeleton-based Action Recognition
\end{IEEEkeywords}

% \fi
%\IEEEpeerreviewmaketitle
\maketitle

% \input{content/01-introduction}
% \input{content/02-related-works}
% \input{content/04-contribution}
% \input{content/05-experiments}
% % \newpage
% \input{content/06-conclusion}

\section{Introduction}\label{sec:introduction}

A human action can be described by a temporal sequence of human body poses, each of which is represented by a set of spatial joint coordinates forming a body skeleton.
Accordingly, skeleton-based action recognition methods process a sequence of skeletons (instead of an image sequence) to recognize the performed action.  
Compared with predicting actions from videos, a sequence of skeleton data not only gives the spatial and temporal features of the body poses, but also provides robustness against different background variations and context noise~\cite{han2017space}. 
The estimation of such skeletal data has become a staple in the human action recognition toolkit thanks to publicly available toolboxes such as OpenPose~\cite{cao2019openpose}.

Early deep learning methods for skeleton-based action recognition either rearrange the body joint coordinates of each skeleton to make a pseudo-image which is used to train a CNN model \cite{kim2017interpretable,liu2017enhanced,naveenkumar2020deep}, or concatenate the human body joints as a sequence of feature vectors and train a RNN model \cite{liu2016spatio,zhang2017view, nikpour2023spatio}. 
However, these methods cannot take advantage of the non-Euclidean structure of the skeletons.
Recently, Graph Convolutional Networks (GCNs) have shown prowess in the modeling of skeleton data \cite{heidari2022gcnChapter}. 
ST-GCN~\cite{yan2018spatial} was the first GCN-based method proposed for skeleton-based action recognition. It uses spatial graph convolutions to extract the per time-step features of each skeleton and employs temporal convolutions to capture time-varying dynamics throughout the skeleton sequence. 
Since its publication, several methods have sprung from ST-GCN, which enhance feature extraction or optimize the structure of the model.

\begin{figure}[t]
    \centering
    \includegraphics[width=1.05\linewidth]{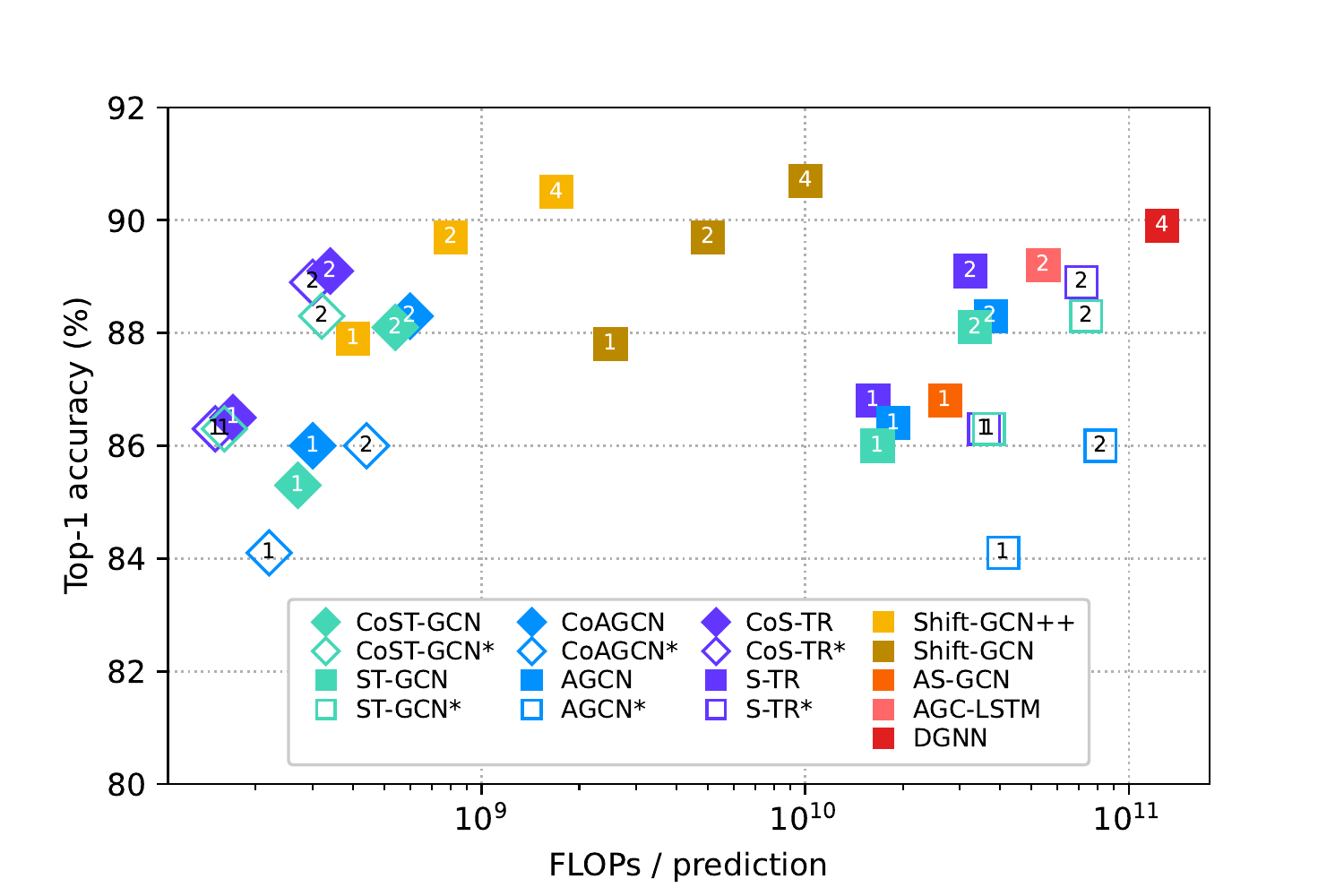}
    \caption{
        \textbf{Accuracy/complexity trade-off} on NTU RGB+D 60 X-Sub for $\sdiamond$ \textit{Continual} and $\ssquare$ prior methods during online inference.
        Numbers denote streams for each method.
        One stream contains the joint modality; two streams add the bone modality; and four streams add joint and bone motion.
        $^*$Architecture modification with stride one and no padding.
    }
    \label{fig:test-acc-vs-flops}
\end{figure}

2s-AGCN~\cite{shi2019two} proposed to learn the graph structure in each GCN layer adaptively based on input graph node similarity and also utilized an attention method which highlights both the existing spatial connections in the graph (bones) and new potential connections between them. 
MS-AAGCN~\cite{shi2020multistream} extended 2s-AGCN by proposing a multi-stream framework which uses four different data streams for training the model. Moreover it enhanced the adaptive graph convolution in 2s-AGCN with a spatio-temporal channel attention module to highlight the most important skeletons, nodes in each skeleton, and features of each node. Following the idea of utilizing different data streams, another multi-stream framework is proposed by ~\cite{li2020learning} which constructs a spatio-temporal view invariant model (STVIM) to capture the spatial and temporal dynamics of joints and bones in skeletons using Geometric Algebra. 
MS-G3D~\cite{liu2020disentangling} has proposed multi-scale graph convolutions for long-range feature extraction, and DGNN~\cite{shi2019skeleton_directed} modeled the spatial connections between the graph nodes with a directed graph and utilized both node features and edge features simultaneously. Similarly, HSR-TSL~\cite{si2020skeleton} is a model with hierarchical spatial reasoning and temporal stack learning network which employs a hierarchical residual graph neural network to capture two-level spatial features, and a temporal stack learning network (TSLN) composed of multiple skip-clip LSTMs to capture temporal dynamics of skeleton sequences. 
%Distinct fine-grained body part features as well as robust temporal dynamics are also captured in some methods~\cite{naveenkumar2020deep}, inspired by ensemble learning, with a composition of several sub-networks.
Recently, STF-Net~\cite{wu2023spatiotemporal} has proposed to capture both robust movement patterns from the skeleton joints and parts topology structures and temporal dependency, using a multi-grain contextual focus module (MCF) and a temporal discrimination focus module (TDF) integrated into a GCN network. 

Unfortunately, the high computational complexity of these GCN-based methods makes them infeasible in real-time applications and resource-constrained online inference settings.
Multiple approaches have been explored to increase the efficiency of skeleton-based action recognition recently:
GCN-NAS~\cite{peng2020learning} and PST-GCN~\cite{heidari2021progressive} are neural architecture search based methods which try to find an optimized ST-GCN architecture to increase the efficiency of the classification task; 
Tripool~\cite{peng2021tripool} is a novel graph pooling method which optimizes a triplet pooling loss to learn an optimized graph topology, by removing the redundant nodes, and learn hierarchical graph representation. 

ShiftGCN~\cite{cheng2020skeleton} replaces graph and temporal convolutions with a zero-FLOPs shift graph operation and point-wise convolutions as an efficient alternative to the feature-propagation rule for GCNs~\cite{kipf2016semi}; 
ShiftGCN++~\cite{cheng2021extremely} boost the efficiency of ShiftGCN further via progressive architecture search, knowledge-distillation, explicit spatial positional encodings, and a Dynamic Shift Graph Convolution; 

SGN~\cite{zhang2020semantics} utilizes semantic information such as joint type and frame index as side information to design a compact semantics-guided neural network (SGN) for capturing both spatial and temporal correlations in joint and frame level; 
TA-GCN~\cite{negarTAGCN} tries to make inference more efficient by selecting a subset of key skeletons, which hold the most important features for action recognition, from a sequence to be processed by the spatio-temporal convolutions.

Yet, none of the above-described GCN-based methods are tailored to online inference, were the input is a continual stream of skeletons and step-by-step predictions are required. 
During online inference, these methods would need to rely on sliding window-based processing, i.e., storing the $T-1$ prior skeletons, appending the newest skeleton to get a sequence of length $T$, and then performing their prediction on the whole sequence. 

In this paper, we reduce such redundant computations by reformulating the ST-GCN and its derived methods as a Continual Inference Network, which processes skeletons one by one and produces updated predictions for each time-step without the need to include past skeletons in every input as is the case for the prior GCN-based methods. 
This is achieved by using Continual Convolutions in place of regular ones for aggregating temporal information, leading to highly reduced number of floating point operations (see Figure \ref{fig:test-acc-vs-flops}). 
In particular, we propose the \textit{Continual} Spatio-Temporal Graph Convolutional Network (\textit{Co}ST-GCN), \textit{Co}AGCN, and \textit{Co}TR-S and evaluate them on the skeleton-based action recognition datasets NTU RGB+D 60~\cite{Shahroudy_2016_NTURGBD}, NTU RGB+D 120~\cite{Liu_2019_NTURGBD120}, and Kinetics Skeleton 400~\cite{kay2017kinetics} with striking results: Our continual models achieve up to 108$\times$ FLOPs reduction, $26\times$ speedup, and $52\%$ reduction in max allocated GPU memory compared to the corresponding non-continual models.

The remainder of the paper is structured as follows: \cref{sec:related-work} provides an introduction to skeleton-based action recognition and of the related methods, from which we derive a continual counterpart, \cref{sec:continual} describes Continual Inference Networks, and \cref{sec:continual-st-gcn} presents our proposed \textit{Continual} Spatio-temporal Graph Convolutional Networks. Experiments on weight transfer strategies, performance benchmarks, and comparisons with prior works are offered in \cref{sec:experiments}, and a conclusion is given in \cref{sec:conclusion}.

%@@@@@@@@@@@@@@@@@@@@@@@@@@@@@@@@@@@@@@@@@@@@@@@@@@@
%@@@@@@@@@@@@@@@@@@@@@@@@@@@@@@@@@@@@@@@@@@@@@@@@@@@

\section{Related Works} \label{sec:related-work}
\subsection{Spatio-Temporal Graph Convolutional Network}\label{sec:gcn}
\begin{figure}[tb]
    \centering
    \includegraphics[width=0.38\linewidth]{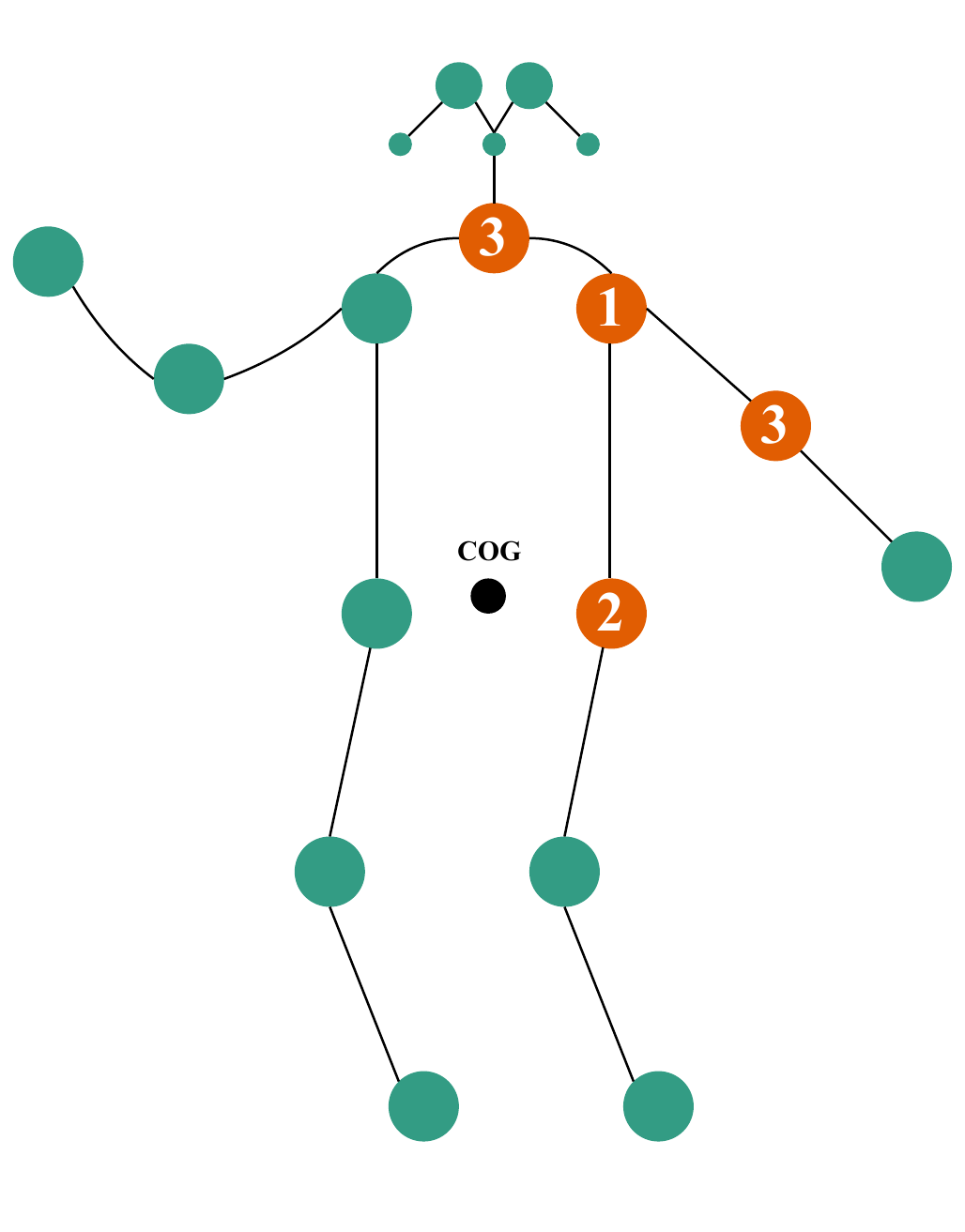}
    \includegraphics[width=0.47\linewidth]{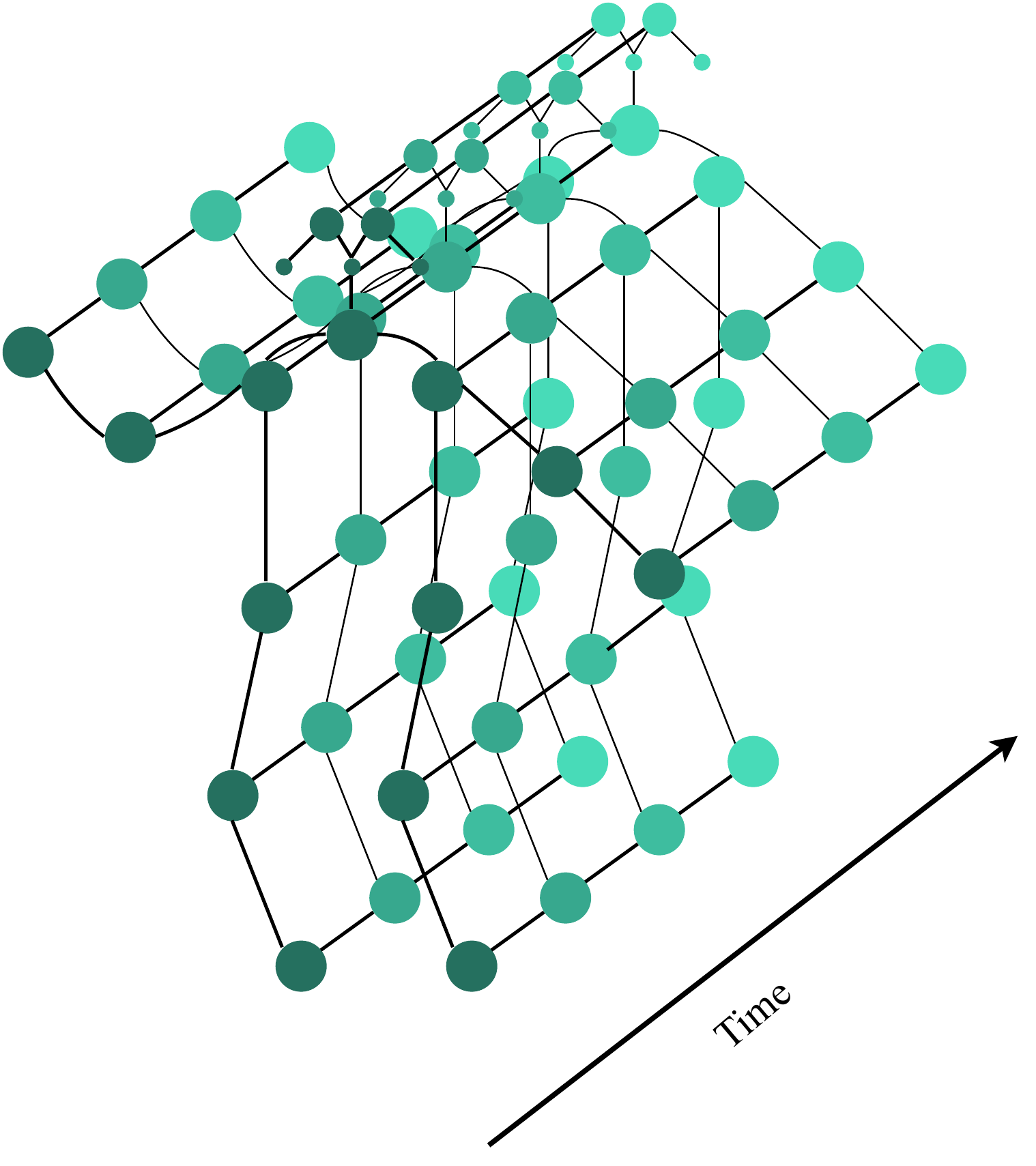}
    \caption{\textbf{Graph illustration} for a spatially partitioned skeleton (left) and spatio-temporal graph (right).}
    \label{fig:ST-Graph-Partitioning}
\end{figure}

GCN-based models for skeleton-based action recognition \cite{yan2018spatial,heidari2021progressive,negarTAGCN} operate on sequences of skeleton graphs. 
The spatio-temporal graph of skeletons $\mathcal{G} = (\mathcal{V}, \mathcal{E})$ has the human body joint coordinates as nodes $\mathcal{V}$ and the spatial and temporal connections between them as edges $\mathcal{E}$. 
Figure~\ref{fig:ST-Graph-Partitioning} (right) illustrates such a spatio-temporal graph where the spatial graph edges encode the human bones and the temporal edges connect the same joints in subsequent time-steps.
We model this graph as a tensor $\mathbf{X} \in \mathbb{R}^{C^{(0)} \times T \times V}$, where $C^{(0)}$ is the number of input-channels of each joint, $T$ denotes the number of skeletons in a sequence, and $V$ is the number of joints in each skeleton. A binary adjacency matrix $\mathbf{A} \in \mathbb{R}^{V \times V}$ encodes the skeleton-structure with ones in positions connecting two vertices in a skeleton and zeros elsewhere.

The ST-GCN~\cite{yan2018spatial} and AGCN~\cite{shi2019two} methods refine the spatial structure of each skeleton by employing a partitioning method which categorizes neighboring nodes of each body joint into three subsets: (1) the root node itself, 
(2) the root's neighboring nodes which are closer to the skeleton's center of gravity (COG) than the root itself, and (3) the remaining neighboring nodes of the root node. An example of this subset partitioning is shown in Figure~\ref{fig:ST-Graph-Partitioning} (left). 
Accordingly, the graph-structure of each skeleton is represented by three normalized binary adjacency matrices $\left \{ \mathbf{A}_p \in \mathbb{R}^{V \times V} \mid  p = 1, 2, 3 \right \}$, each of which is defined as
\begin{equation}
    \hat{\mathbf{A}}_p = \mathbf{D}^{-\frac{1}{2}}_p\mathbf{A}_p\mathbf{D}^{-\frac{1}{2}}_p,  
\end{equation}
where $\mathbf{D}_p$ denotes the degree matrix of the neighboring subset $p$. 
Inspired by the GCN aggregation rule~\cite{kipf2016semi}, the spatial graph convolution receives the hidden representation of the previous layer $\mathbf{H}^{(l-1)}$ as input, where $\mathbf{H}^{(0)} = \mathbf{X}$, and performs the following graph convolution (GC) transformation:
\begin{multline}
    \text{GC}\left(\mathbf{H}^{(l-1)}\right) = \\
    \sigma\left(\text{Res}(\mathbf{H}^{(l-1)}) + \text{BN}\left(\sum_{p} (\mathbf{\hat{A}}_p\otimes \mathbf{M}_p^{(l)})\mathbf{H}^{(l-1)}\mathbf{W}_p^{(l)} \right)\right)
    \label{eq:graph_conv}
\end{multline}
where $\sigma(\cdot)$ denotes a ReLU non-linearity, $\mathbf{W}_p^{(l)} \in \mathbb{R}^{C^{(l)} \times C^{(l-1)}}$ is the weight matrix which transforms the features of the neighboring subset $p$ and $BN(\cdot)$ denotes batch normalization. Moreover, a learnable matrix $\mathbf{M}_p^{(l)} \in \mathbb{R}^{V \times V}$ is multiplied element-wise with its corresponding adjacency matrix $\mathbf{\hat{A}}_p$ as an attention mechanism that highlights the most important connections in each spatial graph. 
In order to retain the model's stability, the input to a layer is added to the transformed features through a residual connection $\text{Res}(\mathbf{H}^{(l-1)})$ which is defined as:
\begin{equation}
    \text{Res}(\mathbf{H}^{(l-1)}) = \begin{cases}\mathbf{H}^{(l-1)}, & C^{(l)} = C^{(l-1),}\\\mathbf{H}^{(l-1)}\mathbf{W}^{(l)}_{res}, & \text{otherwise,}\end{cases}
\end{equation}
where $\mathbf{W}^{(l)}_{res} \in \mathbb{R}^{C^{(l)} \times C^{(l-1)}}$ is a learnable mapping matrix which transforms the layer's input to have the same channel dimension as the layer's output.

The graph convolution block is followed by a temporal convolution, $\text{TC}(\cdot)$, which propagates the features of the graph nodes through different time steps to capture the motions taking place in an action. 
In the temporal graph, each node only has two fixed neighbors which are its corresponding nodes in the previous and next skeletons. The adjacency matrices and partitioning process are not involved in temporal feature propagation.
In practice, the temporal convolution is a standard $2$D convolution which receives the output of the graph convolution obtained in Eq. (\ref{eq:graph_conv}) and performs a transformation with a kernel of size $C^{(l)} \times K \times 1$ to keep the node feature dimension unchanged and aggregate the features through $K$ consecutive time steps. 

The whole spatio-temporal convolution block has the form
\begin{equation}
    \mathbf{H}^{(l)} = \sigma\left(\text{Res}(\mathbf{H}^{(l-1)}) + \text{BN}(\text{TC}(\text{GC}(\mathbf{H}^{(l-1)})))\right).
    \label{eq:spatio-temporal-block}
\end{equation}
The ST-GCN model is composed of multiple such spatio-temporal convolutional blocks. A global average pool and fully connected layer perform the final classification.

\subsection{Adaptive Graph Convolutional Neural Networks}
The fixed graph structure used in \cref{eq:graph_conv} is defined based on natural connections in the human body skeleton which restricts the model's capacity and flexibility in representing different action classes. However, for  some action classes such as ``touching head'' it makes sense to model a connection between hand and head even though such a connection is not naturally present in the skeleton. AGCN~\cite{shi2019two} allows for such possibilities by adopting an adaptive graph convolution which utilizes a data-dependent graph structure as follows:
\begin{multline}
    \text{AGC}\left(\mathbf{H}^{(l-1)}\right) = \\ \sigma\left(\text{Res}(\mathbf{H}^{(l-1)}) + \text{BN}\left(\sum_{p} (\mathbf{\hat{A}}_p + \mathbf{M}_p^{(l)})\mathbf{H}^{(l-1)}\mathbf{W}_p^{(l)} \right)\right), 
    \label{eq:agc}
\end{multline}
where $\mathbf{M}_p^{(l)}$ is defined as:
\begin{equation}
    \mathbf{M}_p^{(l)} = \mathbf{B}_p^{(l)} + \mathbf{C}_p^{(l)}
    \label{eq:agc-m}
\end{equation}

The attention matrix in this definition is composed of two learnable matrices which are optimized along with other model parameters in an end-to-end manner. $\mathbf{B}_p^{(l)} \in \mathbb{R}^{N \times N}$ is a squared matrix that can be unique for each layer and each sample, and $\mathbf{C}_p^{(l)} \in \mathbb{R}^{N \times N}$ is a similarity matrix whose elements determine the strength of the pair-wise connections between nodes. 
This matrix is computed by first transforming the feature matrix $\mathbf{H}^{(l-1)} \in \mathbb{R}^{C^{(l-1)} \times T \times V}$ with two embedding matrices $\mathbf{W}^{(l)}_{p\theta}$, $\mathbf{W}^{(l)}_{p\phi}$ of size $C^{de} \times C^{(l-1)}$. The obtained feature maps are then reshaped to $C^{de}T \times V$ and multiplied to obtain the $\mathbf{C}_p^{(l)} \in \mathbb{R}^{N \times N}$ matrix as follows:
\begin{equation}
    \mathbf{C}_p^{(l)} = \text{softmax}({\mathbf{H}^{(l-1)}}^{\top} \mathbf{W}_{p\theta}^{(l)^{\top}} \mathbf{W}_{p\phi}^{(l)} \mathbf{H}^{(l-1)}),
    \label{eq:agc2}
\end{equation}
where softmax normalizes the matrix values. 
The additive attention mechanism in \cref{eq:agc}, thus, lets the adaptive graph convolution in \cref{eq:agc2} model the skeleton structure as a fully connected graph.

\subsection{Skeleton-based Spatial Transformer Networks}
S-TR~\cite{plizzari2021skeleton} is an attention-based method which 
models dependencies between body joints at each time step using the self-attention operation found in Transformers~\cite{vaswani2017attention}. 
In this method, a Spatial Self-Attention (SSA) module is designed to adaptively learn data-dependent pairwise body joint correlations using multi-head self-attention. 

The SSA module at each layer $l$ applies trainable query, key, and value transformations $\mathbf{W}^{(l)}_q \in \mathbb{R}^{C^{(l-1)} \times dq}$, $\mathbf{W}^{(l)}_k \in \mathbb{R}^{C^{(l-1)} \times dk}$, $\mathbf{W}^{(l)}_v \in \mathbb{R}^{C^{(l-1)} \times dv}$ on the feature vector $\mathbf{h}_i^{t} \in \mathbb{R}^{C^{(l-1)}}$ of node $i$ at time step $t$ to obtain the query, key, and value vectors $\mathbf{q}_i^{t} \in \mathbb{R}^{dq}$, $\mathbf{k}_i^{t} \in \mathbb{R}^{dk}$, $\mathbf{v}_i^{t} \in \mathbb{R}^{dv}$. 
The correlation weight for each pair of $i$, $j$ nodes at time $t$ is obtained using a query-key dot product 
\begin{equation}
    \alpha_{ij}^{t} = {\mathbf{q}_i^{t}}^\top \mathbf{k}_j^{t}. %\forall t \in T, 
    \label{eq:query-key}
\end{equation}
The updated feature vector of node $i$ at time $t$ has size $C^{(l)}$ and is obtained using a weighted feature aggregation of value vectors:
\begin{equation}
    \bar{\mathbf{h}_i^{t}} = \sum_{j} \text{softmax}_j \left( \frac{\alpha_{ij}^t}{\sqrt{dk}}\right)\mathbf{v}_j^t.
    \label{eq:ssa}
\end{equation}

For each attention head, the feature transformation is performed with a different set of learnable parameters while the transformation matrices are shared across all the nodes.
The output features of the SSA module are finally computed by applying a learnable linear transformation on the concatenated features from $S$ attention heads:
\begin{equation}
    \bar{\mathbf{h}_i^{t}} = (\bigparallel_{s=1}^S \bar{\mathbf{h}_{is}^{t}}) \mathbf{W}_o.
    \label{eq:ssa_multihead}
\end{equation}

SSA has similarities to a graph convolution operation on a fully connected graph for which the node connection weights are learned dynamically. 
The first three layers of the S-TR model extract features with GC and TC blocks as defined in Eq. (\ref{eq:spatio-temporal-block}) while in the remaining layers of the model SSA substitutes GC.

\subsection{Continual Inference Networks}\label{sec:continual}
% \section{Continual Convolutional Neural Networks}\label{sec:continual}
First introduced in \cite{hedegaard2021continual} and subsequently formalized in \cite{hedegaard2022cotrans}, Continual Inference Networks are Deep Neural Networks that can operate efficiently on both fixed-size (spatio-)temporal batches of data, where the whole temporal sequence is known up front, as well as on continual data, where new input steps are collected continually and inference needs to be performed efficiently in an online manner for each received frame.
\begin{definition}[\textbf{Continual Inference Network}]
\itshape
A Continual Inference Network is a Deep Neural Network, which
\begin{itemize}
    \item is capable of continual step inference without computational redundancy,
    \item is capable of batch inference corresponding to a non-continual Neural Network,
    \item produces identical outputs for batch inference and step inference given identical receptive fields,
    \item uses one set of trainable parameters for both batch and step inference.
\end{itemize}
\end{definition}
Recurrent Neural Networks (RNNs) are a common family of Deep Neural Networks, which possess the above-described properties. 
3D Convolutional Neural Networks (3D CNNs), Transformers, and Spatio-Temporal Graph Convolutional Networks are not Continual Inference Networks since they cannot make predictions time-step by time-step without considerable computational redundancy; 
% in common computational formulations (using 3D convolutions and Multi-Head Attentions respectively).
they need to cache a sliding window of prior input frames and assemble them into a fixed-size sequence that is subsequently passed through the network to make a new predictions during online inference. 

Recently, \textit{Continual} 3D CNNs were made possible through the proposal of \textit{Continual} 3D Convolutions~\cite{hedegaard2021continual}. 
Likewise, shallow \textit{Continual} Transformers based on \textit{Continual} Dot-product Attentions were introduced in \cite{hedegaard2022cotrans}.
We continue this line of work by extending Spatio-Temporal Graph Convolutional Networks (ST-GCNs) with a \textit{Continual} formulation as well.
To do so, let us first present and expand on the theory on Continual Convolutions.

%@@@@@@@@@@@@@@@@@@@@@@@@@@@@@@@@@@@@@@@@@@@@@@@@@@@@@
%@@@@@@@@@@@@@@@@@@@@@@@@@@@@@@@@@@@@@@@@@@@@@@@@@@@@@

\section{Continual Spatio-Temporal Graph Convolutional Networks}\label{sec:continual-st-gcn-Section}
In the section we present and expand the theory on Continual Convolutions with notes on temporal stride. Then, we describe how the Continual Spatio-Temporal Graph Convolutional Networks are constructed.

\subsection{Continual Convolution} \label{sec:conv}
The Continual Convolution operation produces the exact same output as the regular convolution does, but performs the computation in a streaming fashion while caching intermediary results.

Consider a single channel 2D convolution over an input $\mathbf{X} \in \mathbb{R}^{T \times V}$ with temporal dimension $T$ and a dimension of $V$ vertices.
Given a convolutional kernel with weights $\mathbf{W} \in \mathbb{R}^{K \times V}$, where $K$ is the temporal kernel size, and a bias $w_0$, a regular convolution would compute the output $\mathbf{y}^{(t)}$ for time-step $t \in K..T$ as

\begin{equation}
    \label{eq:conv}
    \mathbf{y}^{(t)} = w_0 + \sum_{k = 1}^{K} \sum_{v = 1}^{V} \mathbf{W}_{k,v} \cdot \mathbf{X}_v^{(t-k-1)}.
\end{equation}
Considering this computation in the context of online processing, where $T \xrightarrow[]{} \infty$ and one input slice $\mathbf{X}^{(t)}$ is revealed in each time step, we find that $K-1$ previous slices, i.e. $(K-1) \cdot V$ values, need to be stored between time-steps.

\begin{figure}[ht]
    \centering
    \includegraphics[width=0.8\linewidth]{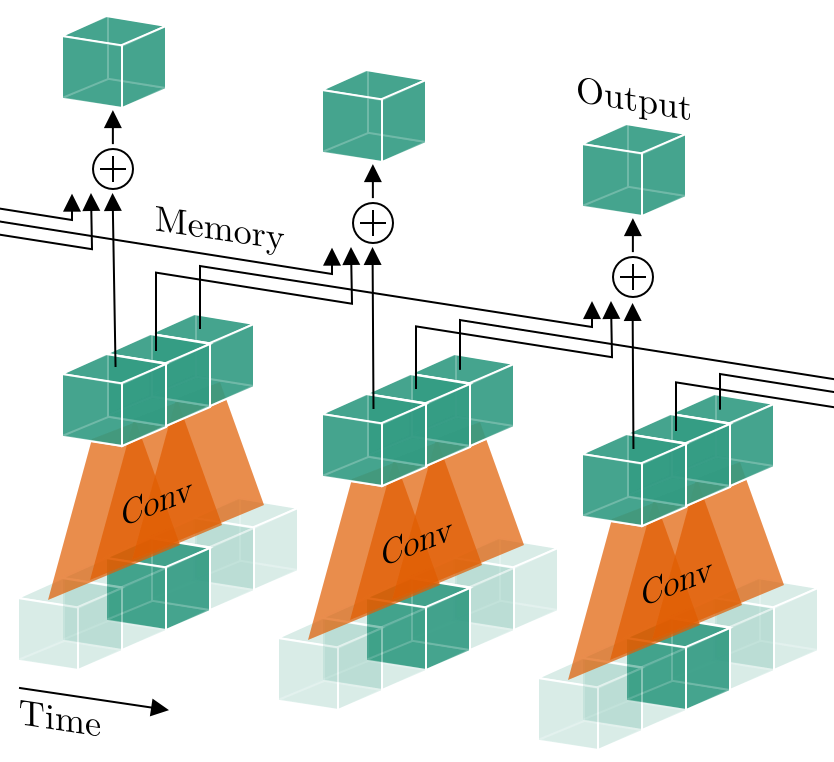}
    \caption{\textbf{Continual Convolutions} are performed in two stages: First, the input is zero-padded and convolved with the convolutional kernel ($K = 3$ in illustration) to produce intermediary results. Subsequently, these are cached and summed up to produce the final output. }
    \label{fig:co-conv}
\end{figure}

An alternative computational sequence is used in Continual Convolutions. 
Here, the input slice $\mathbf{X}^{(t)}$ is convolved with the kernel $\mathbf{W}$ in the same time-step it is received. This is specified in \cref{eq:co-conv-1}. The intermediate results are then cached in memory $\mathbf{m}$ ($K-1$ values stored between time-steps) and aggregated according to \cref{eq:co-conv-2}.

\begin{subequations}\label{eq:co-conv}
    \begin{align}
        \label{eq:co-conv-1}
        \mathbf{m}^{(t)} &= \left[ 
            \sum_{v = 1}^{V} \mathbf{W}_{k,v} \cdot \mathbf{X}_v^{(t)} : \ k \in 1..K
        \right]
        \\
        \label{eq:co-conv-2}
        \mathbf{y}^{(t)} &= w_0 + \sum_{k = 1}^{K} \mathbf{m}^{(t-k-1)}_k
    \end{align}
\end{subequations}
A graphical representation of this is shown in \cref{fig:co-conv}.

\subsection{Delayed Residual}
The temporal convolutions of regular Spatio-Temporal Graph Convolution blocks usually employ zero-padding to ensure equal temporal shape for input and output feature maps.
This zero-padding is discarded for Continual Convolutions to avoid continual redundancies~\cite{hedegaard2021continual}.
To retain weight compatibility between the regular and continual networks, a delay to the residual connection is necessary. This delay amounts to
\begin{equation}
    k_T + (k_T - 1)(d_T-1) - p_T - 1
    \label{eq:delay}
\end{equation}
steps, where $k_T$, $d_T$, and $p_T$ are respectively the temporal kernel size, dilation, and zero-padding of the corresponding regular convolution.

\subsection{Temporal Stride}\label{sec:stride}
\begin{figure}[tb]
    \centering
    \includegraphics[width=1.0\linewidth]{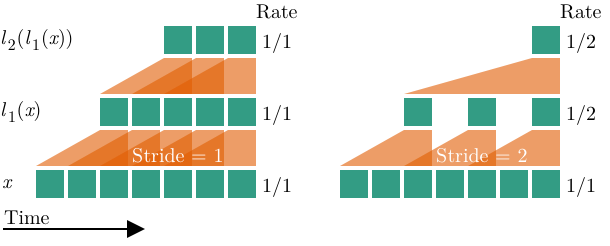}
    \caption{\textbf{Temporal stride} in a Continual Convolution layer $l_1$ with temporal stride larger than one (right) reduces the prediction rate compared to a layer with stride one (left). The rate reduction is inherited by subsequent layers.}
    \label{fig:stride}
\end{figure}

In \cref{sec:conv}, it is assumed that one output is produced for each input received.
However, many spatio-temporal networks including ST-GCN~\cite{yan2018spatial}, AGCN~\cite{shi2019two}, and S-TR~\cite{plizzari2021skeleton}, use temporal stride $>1$ in their temporal convolutions.
For offline computation, this has the beneficial effect of reducing the computational and memory complexity, but in the online computational setting, it also reduces the prediction rate. This is illustrated in \cref{fig:stride}.
For a neural network with $L$ layers, each with a temporal stride $s$, the effective network stride is given by
\begin{equation}
    s_{NN} = \prod_{l=1}^{L} s_l
\end{equation}
and the corresponding network prediction rate is
\begin{equation}
    r_{NN} = 1 / s_{NN}.
\end{equation}
Since a ST-GCN network has two layers with stride two, the corresponding Continual ST-GCN (\textit{Co}ST-GCN) has a prediction rate one fourth the input rate.

\subsection{Continual ST-GCN construction}\label{sec:continual-st-gcn}
Many well-performing methods for skeleton-based action recognition, including the ST-GCN~\cite{yan2018spatial}, AGCN~\cite{shi2019two}, and S-TR~\cite{plizzari2021skeleton}, share a common block structure, which can be described by \cref{eq:spatio-temporal-block}.
Here, the main difference between methods lies in how the graph information is processed, i.e. in their definition of $\text{GC}(\cdot)$. 

The regular skeleton-based methods successively extract complete spatio-temporal skeleton features from the whole sequence with each block before classifying an action.
Considering one block in isolation, the spatio-temporal feature extraction is given by a spatial (graph) convolution followed by a regular temporal convolution. Here, graph convolutions operate locally within a time-step\footnote{AGCN is an exception to this, since the additive attention considers a node's features over all time-steps.},
whereas the temporal convolution does not.
Since the next block $l$ takes as input $\mathbf{H}^{(l-1)}$, the output of the prior block and thereby its temporal convolution, the output of the next spatial (graph) convolution becomes a function of multiple prior time-steps. 
With regular temporal convolutions, features produced by multiple blocks cannot be trivially disentangled and cached in time. Accordingly online operation with per-skeleton predictions can be attained by caching $T-1$ prior skeletons, concatenating these with the newest skeleton, and performing regular spatio-temporal inference. However, this comes with significant computational redundancy, where the complexity of online frame-wise inference is the same as for clip-based inference.

\begin{figure}[tb]
    \centering
    \includegraphics[width=0.6\linewidth]{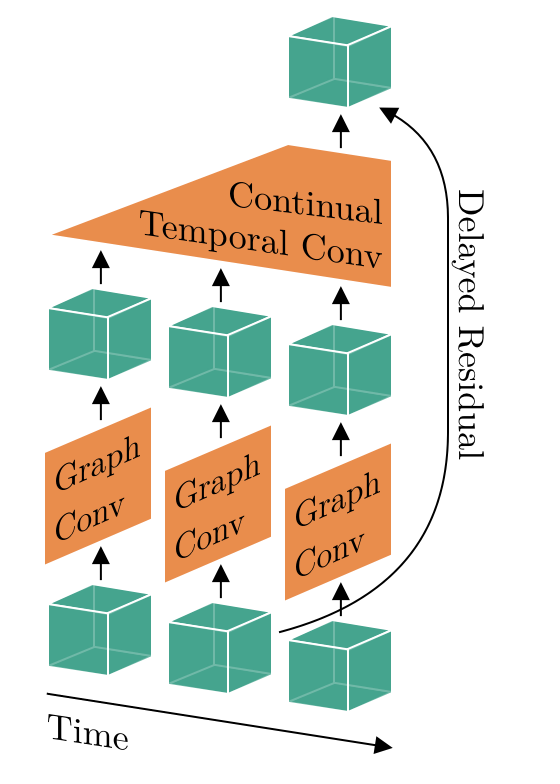}
    \caption{\textbf{Continual Spatio-temporal Graph Convolution Blocks} consist of an in-time Graph Convolution followed by an across-time Continual Convolution (here a kernel size of three is depicted). The residual connection is delayed to ensure temporal alignment with the continual temporal convolution that is weight-compatible with non-continual networks. }
    \label{fig:co-st-gcn-block}
\end{figure}

To alleviate this issue, we propose to employ Continual Convolutions in the temporal modeling of Spatio-temporal Graph Convolutional Networks.
By restricting the $\text{GC}(\cdot)$ function to only operate locally within a time-step% \footnote{For \textit{Co}AGCN, this amounts to calculating separate additive attentions for each time-step from the temporally local note-features, while the spatial operation in \textit{Co}ST-GCN and \textit{Co}S-TR remain the same.}
, we can define a \textit{Continual} Spatio-Temporal block by replacing the original temporal 2D convolution with a continual one. To retain weight-compatibility with regular (non-continual) networks we moreover need to delay the residual to keep temporal alignment.
Given $\mathbf{H}_{l-1}^{(t)}$, i.e. the features of layer $l-1$ in a time-step $t$, the feature in layer $l$ at time $t$ is given by
\begin{equation}
    \mathbf{H}_{l}^{(t)} = \sigma\left(\text{Delay}(\text{Res}(\mathbf{H}_{l-1}^{(t)})) + \text{BN}(Co\text{TC}(\text{GC}(\mathbf{H}_{l-1}^{(t)})))\right).
    \label{eq:co-spatio-temporal-block}
\end{equation}
Here, $\text{Delay}(\text{Res}(\mathbf{H}_{l-1}^{(t)}))$ outputs the delayed residual in a first-in-first-out manner corresponding to the delay of the \textit{Continual} Temporal Convolutional as computed by \cref{eq:delay}.
A graphical illustration of such a block is seen in \cref{fig:co-st-gcn-block}.
It should be noted that the restriction of temporal locality does influence the computations of some skeleton-based action recognition methods. For example, the AGCN originally computes one vertex attention weighting based on the whole spatio-temporal feature-map, whereas a \textit{Continual} AGCN (\textit{Co}AGCN) computes separate vertex attentions for each time-step.

The resulting \textit{Continual} Spatio-temporal Graph Convolutional Network is defined by stacking multiple such blocks\footnote{Following the original ST-GCN, AGCN, and S-TR architectures, ten blocks were used for the networks in this paper.} followed by \textit{Continual} Global Average Pooling~\cite{hedegaard2021continual} and a fully connected layer.
The Continual Inference Networks retain the same computational complexity as regular networks during clip-based inference, but can perform online frame-by-frame predictions much more efficiently, as detailed in \cref{sec:complexity}.
We should note that all methods, which share the the same structure as ST-GCN, i.e. a decoupled temporal and spatial convolution to perform feature transformation and aggregation over the time domain can be transformed to continual version using the approach outlined above.

\subsection{Computational Complexity} \label{sec:complexity}
Denote the time complexity of passing a single skeleton frame through the convolutional blocks with stride 1 by $\mathcal{O}(B)$ and time complexity of utilizing the prediction head by $\mathcal{O}(H)$.
Given an effective clip-size $T$, the complexity of producing a prediction with a regular CNN is approximately $\mathcal{O}(\text{CNN}) \approx T\cdot \mathcal{O}(B) + \mathcal{O}(H)$.
For a Continual CNN, the corresponding complexity is $\mathcal{O}(Co\text{CNN}) \approx \mathcal{O}(B) + \mathcal{O}(H)$. 
Computational savings thus scale linearly with the effective clip-size $T$ and are more prominent the larger $\mathcal{O}(B)$ is compared to $\mathcal{O}(H)$.

\subsection{Limitations}
{
While the computational complexity can be greatly reduced for spatio-temporal GCNs during online processing, i.e., where a prediction is made each time a new skeleton is estimated in a live system, no acceleration occurs during offline processing compared to the original model. When the whole skeleton sequence is available beforehand the inference results and computational complexity are identical to prior works. The benefits of the \textit{Continual} ST-GCN augmentation are thus limited to stream processing for networks which employ temporal convolutions. Accordingly, some networks such as AGCN, whose attention was originally based on the whole spatio-temporal sequence, may need modification to avoid peeking into the future.}

%@@@@@@@@@@@@@@@@@@@@@@@@@@@@@@@@@@@@@@@@@@@@@@@@@@@@
%@@@@@@@@@@@@@@@@@@@@@@@@@@@@@@@@@@@@@@@@@@@@@@@@@@@@

\section{Experiments} \label{sec:experiments}

\subsection{Datasets} \label{sec:datasets}
% We conducted experiments on three widely used large-scale datasets. 
\paragraph{NTU RGB+D 60~\textnormal{\cite{Shahroudy_2016_NTURGBD}}}
A large indoor-captured dataset which is widely used for evaluating skeleton-based action recognition methods. This dataset contains $56{,}880$ action clips and their corresponding $3$D skeleton sequences captured by three Microsoft Kinect-v2 cameras from three different views. The clips are performed by $40$ different subjects and constitute $60$ action classes. The NTU RGB+D 60 dataset comes with two benchmarks, Cross-View (X-View) and Cross-Subject (X-Sub). The X-View benchmark provides $37{,}920$ skeleton sequences coming from the camera views \#2 and \#3 as training data, and $18{,}960$ skeleton sequences coming from the first camera view as test set. The X-Sub benchmark provides $40{,}320$ skeleton sequences from $20$ subjects as training data and $16{,}560$ skeleton sequences from the other $20$ subjects as test data. In this dataset, each skeleton has $25$ body joints with three different channels each, and each action clip comes with a sequence of $300$ skeletons. 

\paragraph{NTU RGB+D 120~\textnormal{\cite{Liu_2019_NTURGBD120}}}
An extension of the NTU RGB+D 60 dataset containing an additional $57{,}600$ skeleton sequences from extra $60$ classes. NTU RGB+D 120 is currently the largest dataset providing $3$D body joint coordinates for skeletons and in total, it contains $114{,}480$ skeleton sequences from $120$ action classes. The action clips in this dataset are performed by $106$ subjects and $32$ different camera setups are used for capturing the videos. This dataset comes with two benchmarks: Cross-Subject (X-Sub) and Cross-Setup (X-Set). The X-Sub benchmark provides the skeleton sequences of $53$ subjects as training data and the remaining skeleton sequences from the other $53$ subjects as test data. 
In the X-Set benchmark, the skeleton sequences with even camera setup IDs are provided as training data and test data contains the remaining skeleton sequences with odd camera setup IDs.

\paragraph{Kinetics Skeleton 400~\textnormal{\cite{kay2017kinetics}}} A widely used dataset for action recognition containing $300{,}000$ video action clips of $400$ different classes which are collected from YouTube. Skeletons were extracted from each frame of these video clips using the OpenPose toolbox~\cite{cao2019openpose}. Each skeleton is represented by $18$ body joints and each body joint contains spatial $2$D coordinates and the estimation confidence score as its three features. We use the dataset version provided by~\cite{yan2018spatial}, which contains $240{,}000$ skeleton sequences as training data and $20{,}000$ skeleton sequences as test data, in our experiments.

\subsection{Experimental Settings}\label{sec:exp-settings}
All models were implemented within the PyTorch framework~\cite{paszke2017automatic} using the Ride library~\cite{hedegaard2021ride}. Models were trained using a SGD optimizer with learning rate $0.1$ at batch size $64$, momentum of $0.9$, and a one-cycle learning rate policy~\cite{smith2019superconvergence} using a cosine annealing strategy. For models which could not fit a batch size of $64$ on a Nvidia RTX 2080 Ti, the learning rate was adjusted following the linear scaling rule~\cite{goyal2017accurate}.
Our source code is available at \url{www.github.com/lukashedegaard/continual-skeletons}.

% The ST-GCN, AGCN and DGNN models are trained on PyTorch deep learning framework \cite{paszke2017automatic} with 4 Nvidia RTX 2080 Ti GPUs, cross entropy loss function and the SGD optimizer with momentum of $0.9$ and weight decay of $0.0001$. 
% The ST-GCN and AGCN models are trained for $50$ epochs with batch size of $64$ on NTU RGB+D 60 dataset and they are trained for $65$ epochs with batch size of $128$ on Kinetics-skeleton dataset. The DGNN model is trained for $120$ epochs with batch size of $32$ on all three datasets. The learning rate is initialized by $0.1$ and it is divided by 10 at epochs $\left [ 30, 40 \right ]$ and $\left [ 45, 55 \right ]$ for NTU RGB+D 60 and kinetics-skeleton datasets, respectively. 
% %it is different for DGNN method 
% %the setting for NTU-120 is also different. 
% {to be continued ...}

\begin{table}[!btp]
\begin{center}
\caption{
    \textbf{Conversion Strategies} from regular (Reg) to Continual (Co) ST-GCN. 
    Noted is the top-1 X-View validation accuracy on NTU RGB+D 60 and the FLOPs per prediction.
    The superscript $p$ and subscript $s$ indicate network padding and stride respectively.
    The arrows $\rightarrow$ and $\xrightarrow[]{FT}$ denote direct conversion and conversion with subsequent fine-tuning.
    Parentheses show the change relative to the baseline with colours indicating \textcolor{lgreen}{improvement} / \textcolor{lred}{deterioration}.
}
\label{tab:conversion}
% \resizebox{\columnwidth}{!}{
\begin{tabular}{lll}
    \toprule
    \textbf{Conversion Strategy}& \textbf{Acc. (\%)}                & \textbf{FLOPs (G)}\\
    \midrule
    % \parbox[t]{1mm}{\multirow{2}{*}{\rotatebox[origin=c]{90}{Clip}}} 
    % \addlinespace[1ex]
    Reg$_{s=4}^{p=\text{eq}}$ (baseline)
                                & 93.4                              & 16.73                                \\
    % \addlinespace[1ex]
    \addlinespace[0.5ex] \cdashline{1-3} \addlinespace[0.5ex]
    Reg$_{s=4}^{p=\text{eq}}$ $\xrightarrow[]{FT}$ Reg$_{s=1}^{p=0}$ %(ST-GCN$^*$) 
                                & 93.8 \phantom{0}\textcolor{lgreen}{($+0.4$)}  & 36.90 \phantom{00}\textcolor{lred}{($\uparrow2.2\times$)} \\
    % \addlinespace[1ex]
    % \parbox[t]{1mm}{\multirow{4}{*}{\rotatebox[origin=c]{90}{Frame}}}     
    % \addlinespace[1ex]
    Reg$_{s=4}^{p=\text{eq}}$ $\rightarrow$ Co$_{s=4}^{p=0}$      
                                & 93.1 \phantom{0}\textcolor{lred}{($-0.3$)}    & \phantom{0}0.27 \phantom{0}\textcolor{lgreen}{($\downarrow63.2\times$)} \\
    % \addlinespace[1ex]
    Reg$_{s=4}^{p=\text{eq}}$ $\rightarrow$ Co$_{s=1}^{p=0}$      
                                & 24.0 \textcolor{lred}{($-69.4$)}   & \phantom{0}0.16 \textcolor{lgreen}{($\downarrow107.7\times$)} \\
    % \addlinespace[1ex]
    Reg$_{s=4}^{p=\text{eq}}$ $\rightarrow$ Co$_{s=1}^{p=0}$ $\xrightarrow[]{FT}$ Co${*}$    
                                & 93.2 \phantom{0}\textcolor{lred}{($-0.2$)}    & \phantom{0}0.16 \textcolor{lgreen}{($\downarrow107.7\times$)} \\
    % \addlinespace[1ex]
    Reg$_{s=4}^{p=\text{eq}}$ $\xrightarrow[]{FT}$ Reg$_{s=1}^{p=0}$ $\rightarrow$ Co${*}$     
                                & 93.8 \phantom{0}\textcolor{lgreen}{($+0.4$)}                              & \phantom{0}0.16 \textcolor{lgreen}{($\downarrow107.7\times$)} \\
    \bottomrule
\end{tabular}
% }
\end{center}
\end{table}

\begin{table*}[!tbp]
\begin{center}
\caption{
    \textbf{NTU RGB+D 60 transfer accuracy and performance benchmarks}. Noted is the top-1 validation accuracy using joints as the only modality. 
    Max mem. is the maximum allocated memory on GPU during inference noted in megabytes.
    Max. mem, FLOPs, and throughput on CPU account for one new prediction with batch size one while throughput on GPU uses the largest fitting power of two as batch size.
    Parentheses indicate the \textcolor{lgreen}{improvement} / \textcolor{lred}{deterioration} relative to the original model.
    % Speed results are the mean $\pm$ std of 100 measurements. 
}
\label{tab:benchmark-speed}
\resizebox{\textwidth}{!}{
\begin{tabular}{lclllllll}
    \toprule
    \multirow{2}{*}{\textbf{Model}} 
    &\textbf{Frames} 
    &\multicolumn{2}{c}{\textbf{Accuracy (\%)}}
    % &\multirow{2}{*}{\textbf{Params (M)}}    
    &\textbf{Params}
    % &\multirow{2}{*}{\textbf{Mem. (MB)}}    
    &\textbf{Max mem.}
    % &\multirow{2}{*}{\textbf{FLOPs (G)}}  
    &\textbf{FLOPs per pred}
    % &\multicolumn{4}{c}{\textbf{Speed (evaluations/s)}}
    &\multicolumn{2}{c}{\textbf{Throughput (preds/s)}}
        \\ 
            \cline{3-4} 
            \cline{8-9}
        & \textbf{per pred} & X-Sub & X-View & (M) & (MB) & (G) & CPU & GPU
    \\
    \midrule
    % \multirow{2}{*}{Clip}                   
    %                                   Fr/p    X-Sub   X-View  Params (M)  Mem. (MB)              FLOPS(G)    CPU                  GPU
    ST-GCN                              & 300   & 86.0  & 93.4  & 3.14      & \phantom{0}45.3    & 16.73     & \phantom{0}2.3     & \phantom{0}180.8 \\
    
    ST-GCN$^*$                          & 300   & 86.3 \textcolor{lgreen}{($+0.3$)}  & 93.8 \textcolor{lgreen}{($+0.4$)}  & 3.14      & \phantom{0}72.6 \textcolor{lred}{($160\%$)}   & 36.90 \textcolor{lred}{\phantom{00}($\uparrow2.2\times$)}      & \phantom{0}1.1 \textcolor{lred}{\phantom{0}($\downarrow2.1\times$)}     & \phantom{00}90.4 \phantom{0}\textcolor{lred}{($\downarrow2.0\times$)} \\
    % \multirow{2}{*}{Frame}
    
    \textit{Co}ST-GCN                   & 4     & 85.3 \textcolor{lred}{($-0.7$)}  & 93.1 \textcolor{lred}{($-0.3$)}   & 3.14       & \phantom{0}36.0 \phantom{0}\textcolor{lgreen}{($79\%$)}     & \phantom{0}0.27 \textcolor{lgreen}{\phantom{0}($\downarrow63.2\times$)}      & 32.3 \textcolor{lgreen}{($\uparrow14.0\times$)}     & 2375.2 \textcolor{lgreen}{($\uparrow13.1\times$)} \\
    
    \textit{Co}ST-GCN$^*$               & 1     & 86.3 \textcolor{lgreen}{($+0.3$)} & 93.8 \textcolor{lgreen}{($+0.4$)}   & 3.14     & \phantom{0}36.1 \phantom{0}\textcolor{lgreen}{($80\%$)}    & \phantom{0}0.16 \textcolor{lgreen}{ ($\downarrow107.7\times$)}     & 46.1 \textcolor{lgreen}{($\uparrow20.0\times$)}     & 4202.2 \textcolor{lgreen}{($\uparrow23.2\times$)} \\
    
    \midrule
    AGCN                                & 300   & 86.4  & 94.3  & 3.47   & \phantom{0}48.4  & 18.69     & \phantom{0}2.1   & \phantom{0}146.2 \\
    
    AGCN$^*$                            & 300   & 84.1 \textcolor{lred}{($-2.3$)}  & 92.6 \textcolor{lred}{($-1.7$)}  & 3.47        & \phantom{0}76.4 \textcolor{lred}{($158\%$)}   & 40.87
    \textcolor{lred}{\phantom{00}($\uparrow2.2\times$)}     & \phantom{0}1.0 \textcolor{lred}{\phantom{0}($\downarrow2.1\times$)}     & \phantom{00}71.2 \phantom{0}\textcolor{lred}{($\downarrow2.0\times$)} \\
    
    \textit{Co}AGCN                     & 4     & 86.0 \textcolor{lred}{($-0.4$)} & 93.9 \textcolor{lred}{($-0.4$)} & 3.47     & \phantom{0}37.3 \phantom{0}\textcolor{lgreen}{($77\%$)}       & \phantom{0}0.30 \textcolor{lgreen}{\phantom{0}($\downarrow63.2\times$)}     & 25.0 \textcolor{lgreen}{($\uparrow11.9\times$)}     & 2248.4 \textcolor{lgreen}{($\uparrow15.4\times$)} \\
    
    \textit{Co}AGCN$^*$                 & 1     & 84.1 \textcolor{lred}{($-2.3$)}  & 92.6 \textcolor{lred}{($-1.7$)}  & 3.47   & \phantom{0}37.4 \phantom{0}\textcolor{lgreen}{($77\%$)}       & \phantom{0}0.17 \textcolor{lgreen}{($\downarrow108.8\times$)}      & 30.4 \textcolor{lgreen}{($\uparrow14.5\times$)}     & 3817.2 \textcolor{lgreen}{($\uparrow26.1\times$)} \\
    
    \midrule
    S-TR                                & 300   & 86.8     & 93.8     & 3.09    & \phantom{0}74.2      & 16.14         & \phantom{0}1.7     & \phantom{0}156.3 \\
    
    S-TR$^*$                            & 300   & 86.3 \textcolor{lred}{($-0.5$)}      & 92.4 \textcolor{lred}{($-1.4$)}     & 3.09    & 111.5 \textcolor{lred}{($150\%$)}      & 35.65 \textcolor{lred}{\phantom{00}($\uparrow2.2\times$)}        & \phantom{0}0.8 \textcolor{lred}{\phantom{0}($\downarrow2.1\times$)}     & \phantom{00}85.1 \phantom{0}\textcolor{lred}{($\downarrow1.8\times$)} \\
    
    \textit{Co}S-TR                     & 4     & 86.5 \textcolor{lred}{($-0.3$)}     & 93.3 \textcolor{lred}{($-0.5$)}      & 3.09    & \phantom{0}35.9 \phantom{0}\textcolor{lgreen}{($48\%$)}      & \phantom{0}0.22 \textcolor{lgreen}{\phantom{0}($\downarrow63.2\times$)}         & 30.3 \textcolor{lgreen}{($\uparrow17.8\times$)}     & 2189.5 \textcolor{lgreen}{($\uparrow14.0\times$)} \\
    
    \textit{Co}S-TR$^*$                 & 1     & 86.3 \textcolor{lred}{($-0.3$)}     & 92.4 \textcolor{lred}{($-1.4$)}     & 3.09   & \phantom{0}36.1 \phantom{0}\textcolor{lgreen}{($49\%$)}       & \phantom{0}0.15 \textcolor{lgreen}{($\downarrow107.6\times$)}         & 43.8 \textcolor{lgreen}{($\uparrow25.8\times$)}     & 3775.3 \textcolor{lgreen}{($\uparrow24.2\times$)} \\
    
    \bottomrule
\end{tabular}
}
\end{center}
\end{table*}

\subsection{Conversion and Fine-tuning Strategies}\label{sec:ft-strategies}
Though regular and Continual CNNs are weight-compatible, the direct transfer of weights is imperfect if the regular CNN was trained with zero-padding~\cite{hedegaard2021continual}.
As in most CNNs, it is common practice to utilize padding in skeleton-based spatio-temporal networks to retain the temporal feature size in consecutive layers (though temporal shrinkage is not a concern given the long input clips).

Another common design choice, which has a significant impact in on the performance of Continual Inference Networks, is the utilization of temporal stride larger than one. 
For regular networks, this has the benefit of reducing the computational complexity per clip prediction. 
In Continual Inference Networks, however, it reduces the prediction rate, and actually increases the complexity per prediction (see \cref{sec:stride}).
In the continual case, it would thus be computationally beneficial to reduce the stride of all layers to one. However, this results in a stride-inflicted \textit{model-shift}.

\begin{figure}
    \centering
    \includegraphics[width=\linewidth]{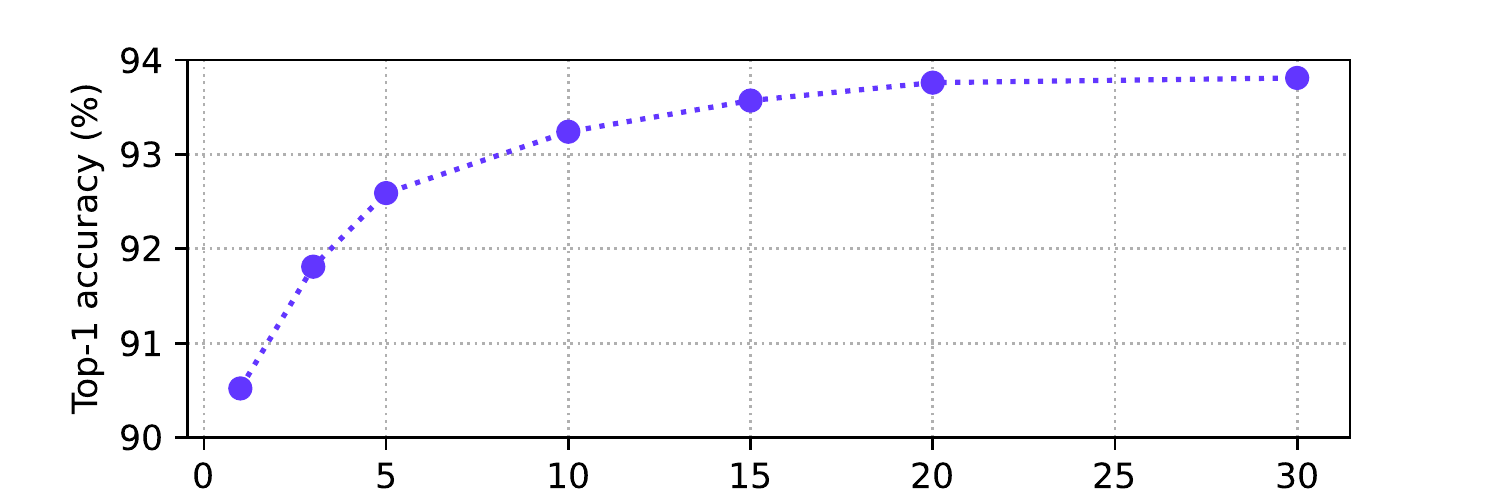}
    \caption{\textbf{Fine-tuning epochs} and associated top-1 accuracy on NTU RGB+D 60 X-View for a transfer from a pre-trained ST-GCN with zero-padding and accumulated stride of four to an equivalent (\textit{Co})ST-GCN$^*$ with no zero-padding and stride one. }
    \label{fig:tune-num-epochs}
\end{figure}

Thus far, the \textit{model-shift} inflicted by padding removal and stride reduction, as well as how to best perform the conversion from a regular CNN to a Continual CNN in such cases has not been studied. 
In this set of experiments, we explore strategies on how to best convert and fine-tune regular networks to achieve good frame-by-frame performance.
We use a standard ST-GCN~\cite{yan2018spatial} trained on joints only as our starting-point, and explore the accuracy achieved by:
\begin{enumerate}
    \itemsep0em 
    \item Converting to from regular network with equal padding and stride four (Reg$_{s=4}^{p=\text{eq}}$) to a Continual Inference Network, where zero-padding is omitted (Co$_{s=4}^{p=0}$).
    \item Reducing the network stride to one without fine-tuning (Co$_{s=1}^{p=0}$).
    \item Fine-tuning the Co$_{s=1}^{p=0}$ network (= Co${*}$).
    \item Fine-tuning a conversion-optimal regular network which has no zero-padding and a stride of one (Reg$_{s=1}^{p=0}$).
    \item Converting from Reg$_{s=1}^{p=0}$ to Continual (= Co${*}$).
\end{enumerate}
As seen in \cref{tab:conversion}, the direct transfer of weights was found to have a modest negative impact on the accuracy (by $-0.3\%$) due the removal of zero-padding. This is considerably less than was found in \cite{hedegaard2021continual}. Our conjecture is that the smaller amount of zeros relative to clip size used in skeleton-based recognition (8 zeros per 300 frames or 2.67\%) compared to video-based recognition (e.g., 2 zeros per 16 frames or or 12.5\%) makes the removal of zero-padding less detrimental since zeros contribute relatively less to the downstream features.
Lowering the stride to one and removing zero-padding reduced accuracy by a substantial amount but allowed the Continual Inference Network to operate at much lower FLOPs. % ($107.7\times$ reduction for stride 1 compared with $63.2\times$ reduction for accumulated stride 4).
This accuracy drop is alleviated equally effectively by either (a) initializing the Co$_{s=1}^{p=0}$ with standard weights and fine-tuning in the continual regime or (b) first fine-tuning the conversion-optimal regular network (Reg$_{s=1}^{p=0}$) and subsequently converting to a Continual Inference Network, though the latter had lower training times in practice.
We fine-tuned the networks using the settings described in \cref{sec:exp-settings}.
As visualised in \cref{fig:tune-num-epochs}, we found 20 epochs of fine-tuning using the settings described in \cref{sec:exp-settings} recover accuracy on NTU RGB+D 60 with additional training yielding only marginal differences. 
Following this approach the (padding zero, stride one) optimized Continual ST-GCN (\textit{Co}ST-GCN$^*$) achieves a similar prediction accuracy while reducing the computational complexity by a factor $107.7\times$ relative to original ST-GCN!

\subsection{Conversion of Attention Architectures}
As we explored in \cref{sec:ft-strategies}, the ST-GCN network architecture can easily be modified and fine-tuned to achieve high accuracy for frame-by-frame predictions with exceptionally low computational complexity.
A natural follow-up question is whether this conversion is equally successful for more complicated spatio-temporal architectures that employ attention mechanisms.
To investigate this, we conduct a similar transfer for two recent ST-GCN variants,
the Adaptive GCN (AGCN)~\cite{shi2019two} and the Spatial Transformer Network (S-TR)~\cite{plizzari2021skeleton}.
While S-TR is easily converted to a Continual Inference Network (\textit{Co}S-TR) by replacing convolutions, residuals and pooling operators with Continual ones, the AGCN requires additional care.
In the original version of AGCN, the vertex attention matrix $\mathbf{C}_p$ (see \cref{eq:agc2}) is computed from the global representations in the layer over all time-steps. 
Since this operation would be acausal in the context of a Continual Inference Network, we restrict it to utilize only the frame-specific subset of features. 
As a fine-tuning strategy, we first make the conversion from regular network to a conversion-optimal network, and subsequently convert and evaluate the continual version.

Our results are presented in \cref{tab:benchmark-speed}.
Here we see that all three architectures can be successfully converted to continual versions. 
The fine-tuned conversion-optimal models (marked by $^*$) generally exhibit a higher computational complexity than their source models due to their stride decrease. While the ST-GCN$^*$ attained increased performance by lowering stride, AGCN$^*$ and S-TR$^*$ suffer slight accuracy deterioration. This may be due to smaller receptive fields of their attention mechanisms, which likely benefit from observing a larger context.
Unlike the transfer from the original models with padding and stride four to continual models, the continual models with weights from ST-GCN$^*$, AGCN$^*$, and S-TR$^*$, i.e. \textit{Co}ST-GCN$^*$, \textit{Co}AGCN$^*$, and \textit{Co}S-TR$^*$ attain the exact same accuracy as their source models on both the X-Sub and X-View benchmarks, with two orders of magnitude less FLOPs per prediction during online inference.

\subsection{Speed and Memory}
Diving deeper into the differences between regular and continual networks, we conduct throughput benchmarks on a MacBook Pro 16” with a 2.6 GHz 6-Core Intel Core i7 CPU and a NVIDIA RTX 2080 Ti GPU.
Here, we measure the prediction-time as the time it takes to transfer an input of batch size one from CPU to GPU (if applicable), perform inference, and transfer the results back to CPU again.
On CPU, a batch size of one is used, while for GPU, the largest fitting power of two is employed (i.e. \{128, 64, 256, 256\} for the \{Reg, Reg$^*$, \textit{Co}, and \textit{Co}$^*$\} models).
We measure the maximum allocated memory during inference on GPU for batch size one.

As seen in \cref{tab:benchmark-speed}, the change in speed relative to the original models follow a similar trend to those seen for FLOPs. The non-continual stride one variants (denoted by $^*$) exhibit roughly half the speed of the original models, while the continual models enjoy more than a magnitude speed up on both CPU and GPU. As expected, the continual stride one models (\textit{Co}$^*$) attain the largest inference throughput. These relative speed-ups are lower than the relative FLOPs reductions due to the read/writes of internal intermediary features in the Continual Convolutions since these are not accounted for by the FLOPs metric while still adding to the runtime. This gap could be reduced on hardware with in- or near-memory computing.

Considering the maximum allocated memory at inference, 
we find that the continual models reduce memory by 20-52\%. While the Continual Convolution and -Pooling layers do add some internal state that adds to the memory consumption, the intermediary features that are passed between network layers are much smaller, i.e. one frame instead of 75 to 300 frames.

\begin{table}[!btp]
\begin{center}
\caption{
    \textbf{NTU RGB+D 60} comparison with recent methods, grouped by clip- and frame-based inference.
    Noted are the number of streams (S.), top-1 validation accuracy, and FLOPs per prediction.
    $^\dagger$Results for our implementation. Highlights indicate \textbf{best}, \textbf{\textit{next-best}} and \colorbox{\paretocolor}{Pareto optimal} results.
}
\label{tab:benchmark-ntu60}
\resizebox{\columnwidth}{!}{
\begin{tabular}{llcccr}
    \toprule
    &\multirow{2}{*}{\textbf{Model}}
    &\multirow{2}{*}{\textbf{S.}} 
    &\multicolumn{2}{c}{\textbf{Accuracy (\%)}}
    % &\multirow{2}{*}{\textbf{Params (M)}}   
    &\textbf{FLOPs}
    % &\multirow{2}{*}{\textbf{FLOPs (G)}}  
        \\ \cline{4-5} 
        &&& X-Sub & X-View & (G)
    \\
    \midrule
    
    Clip

    & SGN~\cite{zhang2020semantics}                 & 1             & 89.4  & 94.5    & - \\ 
    % PR 
    & STVIM~\cite{li2020learning}                   & 4 
        & 85.6 & 92.0     & - \\
    % PR 
    & HSR-TSL~\cite{si2020skeleton}                 & 2 
        & 87.7 & 94.4 & - \\
    & MS-G3D~\cite{liu2020disentangling}            & 1             & 89.4  & 95.0    & - \\ 
    &                                               & 2             & \textbf{91.5}  & 96.2    & - \\ 
    & ST-TR~\cite{plizzari2021skeleton}             & 1             & 89.2  & 95.8    & - \\ 
    &                                               & 2             & 90.3  & 96.3    & - \\ 
    & MS-AAGCN~\cite{shi2020multistream}            & 4             & 90.0  & 96.2    & -   \\  
    & DGNN~\cite{shi2019skeleton_directed}          & 4             & 89.9  & 96.1    & 126.80 \\ 
    & AS-GCN~\cite{li2019actional}                  & 1             & 86.8  & 94.2    & 27.00 \\ 
    & AGC-LSTM~\cite{si2019attention}               & 2             & 89.2  & 95.0    & 54.40 \\ 
    
    % PR 
    & Tripool~\cite{peng2021tripool}                & 1             & 88.0  & 95.3    & 11.76 \\
    &                                               & 2             & 89.5  & 96.4    & - \\
    &                                               & 3             & 90.0  & \textbf{96.7} & - \\ 

    & {STF-Net}~\cite{wu2023spatiotemporal}      & {1}             & {88.8}  & {95.0}    & {7.5} \\
    &                                               & {2}             & {90.8}  & {96.2}    & {-} \\
    &                                               & {4}             & \textbf{\textit{{91.1}}}  & \textbf{\textit{{96.5}}}    & - \\

    & {STH-DRL}~\cite{nikpour2023spatio}         & {1}        & {90.8}  & \textbf{{96.7}}    & {-} \\

    & ShiftGCN~\cite{cheng2020skeleton}             & 1             & 87.8  & 95.1    & 2.50 \\ 
    &                                               & 2             & 89.7  & 96.0    & 5.00 \\ 
    &                                               & 4             & \colorbox{\paretocolor}{90.7}  & \colorbox{\paretocolor}{\textbf{\textit{96.5}}}    & \colorbox{\paretocolor}{10.00} \\ 
    & ShiftGCN++~\cite{cheng2021extremely}          & 1             & \colorbox{\paretocolor}{87.9}  & \colorbox{\paretocolor}{94.8}    & \colorbox{\paretocolor}{\textbf{0.40}} \\ 
    &                                               & 2             & \colorbox{\paretocolor}{89.7}  &  \colorbox{\paretocolor}{95.7}    & \colorbox{\paretocolor}{\textbf{\textit{0.80}}} \\ 
    &                                               & 4             &  \colorbox{\paretocolor}{90.5}  & \colorbox{\paretocolor}{96.3}    & \colorbox{\paretocolor}{1.70} \\ 
    \addlinespace[0.5ex] \cline{2-6} \addlinespace[0.5ex]
    
    & ST-GCN$^\dagger$                              & 1             & 86.0  & 93.4    & 16.73 \\
    &                                               & 2             & 88.1  & 94.9    & 33.46 \\
  
    & AGCN$^\dagger$                                & 1             & 86.4  & 94.3    & 18.69 \\
    &                                               & 2             & 88.3  & 95.3    & 37.38 \\
  
    & S-TR$^\dagger$                                & 1             & 86.8  & 93.8    & 16.20 \\ 
    &                                               & 2             & 89.1  & 95.3    & 32.40 \\
    
    \midrule
    
    Frame
    & Deep-LSTM~\cite{Shahroudy_2016_NTURGBD}       & 1             & 60.7  & 67.3    & - \\ 
    & VA-LSTM~\cite{zhang2017view}                  & 1             & 79.2  & 87.7    & - \\ 
    \addlinespace[0.5ex] \cline{2-6} \addlinespace[0.5ex]
    & \textit{Co}ST-GCN (ours)                      & 1             & 86.0  & 93.4    & 0.27 \\
    &                                               & 2             & 88.1  & 94.8    & 0.54 \\
    & \textit{Co}ST-GCN$^*$ (ours)                  & 1             & 86.3  & \colorbox{\paretocolor}{93.8}    & \colorbox{\paretocolor}{\textbf{\textit{0.16}}} td\\
    &                                               & 2             & 88.3  & \colorbox{\paretocolor}{\textbf{\textit{95.0}}}    &  \colorbox{\paretocolor}{0.32} \\
    & \textit{Co}AGCN (ours)                        & 1             & 86.4  & 94.2    & 0.30 \\
    &                                               & 2             & 88.2  & \textbf{95.3}    & 0.60 \\
    & \textit{Co}AGCN$^*$ (ours)                    & 1             & 84.1  & 92.6    & 0.22 \\
    &                                               & 2             & 86.0  & 93.1    & 0.44 \\
    & \textit{Co}S-TR (ours)                        & 1             & \colorbox{\paretocolor}{86.5}  & 93.5    & \colorbox{\paretocolor}{0.17} \\
    &                                               & 2             & \colorbox{\paretocolor}{\textbf{88.8}}  & \colorbox{\paretocolor}{\textbf{95.3}}    & \colorbox{\paretocolor}{0.34} \\
    & \textit{Co}S-TR$^*$ (ours)                    & 1             & \colorbox{\paretocolor}{86.3}  & \colorbox{\paretocolor}{92.4}    &  \colorbox{\paretocolor}{\textbf{0.15}} \\
    &                                               & 2             & \colorbox{\paretocolor}{\textbf{\textit{88.9}}}  & \colorbox{\paretocolor}{94.8}    & \colorbox{\paretocolor}{0.30} \\
    \bottomrule
\end{tabular}
}
\end{center}
% \vspace{-10pt}
\end{table}

\begin{table}[!tbp]
\begin{center}
\caption{
    \textbf{NTU RGB+D 120} comparison with recent methods, grouped by clip- and frame-based inference.
    Noted are the number of streams (S.), top-1 validation accuracy, and FLOPs per prediction.
    $^\dagger$Results for our implementation.
    Highlights indicate \textbf{best}, \textbf{\textit{next-best}} and \colorbox{\paretocolor}{Pareto optimal} results.
}
\label{tab:benchmark-ntu-120}
\resizebox{\columnwidth}{!}{
\begin{tabular}{llcccr}
    \toprule
    &\multirow{2}{*}{\textbf{Model}}
    &\multirow{2}{*}{\textbf{S.}} 
    &\multicolumn{2}{c}{\textbf{Accuracy (\%)}}
    &\textbf{FLOPs}
        \\ \cline{4-5} 
        &&& X-Sub & X-Set & (G)
    \\
    \midrule
    % \parbox[t]{1mm}{\multirow{7}{*}{\rotatebox[origin=c]{90}{Clip}}}                    
    Clip      
    %                                               Streams     X-Sub       X-Set   FLOPs
    % & Part-Aware LSTM~\cite{liu2019ntu}             & 1         & 25.5      & 26.3  & - \\
    & ST-LSTM~\cite{liu2016spatio}                  & 1         & 55.7      & 57.9  & - \\
    % & TSRJI~\cite{caetano2019skeleton}              & 1         & 67.9      & 62.8  & - \\
    & SGN~\cite{zhang2020semantics}                 & 1         & 79.2      & 81.5  & - \\
    & MS-G3D~\cite{liu2020disentangling}            & 2         & \textbf{86.9}      & \textbf{88.4}  & - \\
    % PR 
    & Tripool~\cite{peng2021tripool}                & 1         & 80.1      & 82.8  & - \\
    
    & {STF-Net}~\cite{wu2023spatiotemporal}      & {4}         & \textbf{\textit{{86.5}}}      & \textbf{\textit{{88.2}}}  & {7.5} \\

    & ShiftGCN~\cite{cheng2020skeleton}             & 1         & 80.9      & 83.2  & 2.50 \\
    &                                               & 2         & 85.3      & 86.6  & 5.00 \\
    &                                               & 4         & 85.9      & 87.6  &  10.00 \\
    & ShiftGCN++~\cite{cheng2021extremely}          & 1         & \colorbox{\paretocolor}{80.5}     &  \colorbox{\paretocolor}{83.0}  & \colorbox{\paretocolor}{\textbf{0.40}} \\
    &                                               & 2         & \colorbox{\paretocolor}{84.9}      & \colorbox{\paretocolor}{86.2}  & \colorbox{\paretocolor}{\textbf{\textit{0.80}}} \\
    &                                               & 4         & \colorbox{\paretocolor}{85.6}      & \colorbox{\paretocolor}{87.2}  & \colorbox{\paretocolor}{1.70} \\
    \addlinespace[0.5ex] \cline{2-6} \addlinespace[0.5ex]
    & ST-GCN$^\dagger$                              & 1         & 79.0      & 80.7  & 16.73 \\
    &                                               & 2         & 83.7      & 85.1  & 33.46 \\
  
    & AGCN$^\dagger$                                & 1         & 79.7      & 80.7  & 18.69 \\
    &                                               & 2         & 84.0      & 85.4  & 37.38 \\
  
    & S-TR$^\dagger$                                & 1         & 80.2      & 81.8  & 16.20 \\ 
    &                                               & 2         & 84.8      & 86.2  & 32.40 \\

    \midrule
    Frame 
    & \textit{Co}ST-GCN (ours)                      & 1         & 78.9      & 80.7     & 0.27 \\
    &                                               & 2         & 83.7      & 85.1     & 0.54 \\
    & \textit{Co}ST-GCN$^*$ (ours)                  & 1         & 79.4      & 81.6     & \textbf{\textit{0.16}} \\
    &                                               & 2         & 84.0      & 85.5     & 0.32 \\
    & \textit{Co}AGCN (ours)                        & 1         & 79.6      & 80.7     & 0.30 \\
    &                                               & 2         & 84.0      & 85.3     & 0.60 \\
    & \textit{Co}AGCN$^*$ (ours)                    & 1         & 77.3      & 79.1     & 0.22 \\
    &                                               & 2         & 80.4      & 82.0     & 0.44 \\
    & \textit{Co}S-TR (ours)                        & 1         & \colorbox{\paretocolor}{80.1}      & 81.7     & \colorbox{\paretocolor}{0.17} \\
    &                                               & 2         & \textbf{\textit{84.8}}      & \colorbox{\paretocolor}{\textbf{86.1}}     & \colorbox{\paretocolor}{0.34} \\
    & \textit{Co}S-TR$^*$ (ours)                    & 1         & \colorbox{\paretocolor}{79.7}      & \colorbox{\paretocolor}{81.7}     & \colorbox{\paretocolor}{\textbf{0.15}} \\
    &                                               & 2         & \colorbox{\paretocolor}{\textbf{84.8}}      & \colorbox{\paretocolor}{\textbf{\textit{86.1}}}     & \colorbox{\paretocolor}{0.30} \\
    \bottomrule
\end{tabular}
}
\end{center}
% \vspace{-10pt}
\end{table}

\begin{table}[!tbp]
\begin{center}
\caption{
    \textbf{Kinetics Skeleton 400} comparison with recent methods, grouped by clip- and frame-based inference.
    Noted are the number of streams (S.), top-1 and top-5 validation accuracy, and FLOPs per prediction.
    $^\dagger$Results for our implementation.
    Highlights indicate \textbf{best}, \textbf{\textit{next-best}} and \colorbox{\paretocolor}{Pareto optimal} results.
}
\label{tab:benchmark-kinetics}
\resizebox{\columnwidth}{!}{
\begin{tabular}{llcccr}
    \toprule
    &\multirow{2}{*}{\textbf{Model}} 
    &\multirow{2}{*}{\textbf{S.}} 
    &\multicolumn{2}{c}{\textbf{Accuracy (\%)}}
    &\textbf{FLOPs}
    \\ \cline{4-5}
        &&& Top-1 & Top-5 & (G)
    \\
    \midrule
    Clip
    
    % & Feature Enc.~\cite{fernando2015modelling, yan2018spatial} & 1     & 14.9      & 25.8  & - \\
    & Deep LSTM~\cite{shahroudy2016ntu,yan2018spatial}          & 1     & 16.4      & 35.3  & - \\
    & TCN~\cite{kim2017interpretable,yan2018spatial}            & 1     & 20.3      & 40.0  & - \\
    & AS-GCN~\cite{li2019actional}                              & 1     & 34.8      & 56.5  & - \\
    & ST-GR~\cite{li2019spatio}                                 & 1     & 33.6      & 56.1  & - \\
    % PR 
    & Tripool~\cite{peng2021tripool}                            & 1     & 34.1      & 56.2  & - \\
    & {STF-Net}~\cite{wu2023spatiotemporal}                  & {4}     & {36.1}      & {58.9}  & {-}\\

    & DGNN~\cite{shi2019skeleton_directed}                      & 4     & \textbf{\textit{36.9}}      & 59.6  & - \\
    & MS-G3D~\cite{liu2020disentangling}                        & 2     & \textbf{38.0}      & \textbf{\textit{60.9}}  & - \\
    & MS-AAGCN~\cite{shi2020multistream}                        & 4     & 37.8      & \textbf{61.0}  & - \\  
    % NN
    % ED-GCN~\cite{?} & 2 & 36.9, 59.0 & - \\
    % TIP
    %& Hyper-GNN~\cite{hao2021hypergraph}                        & 3     & 37.1      & 60.0  & - \\  % TIP 
    \addlinespace[0.5ex] \cline{2-6} \addlinespace[0.5ex]
    & ST-GCN$^\dagger$                                          & 1     & \colorbox{\paretocolor}{33.4}      & \colorbox{\paretocolor}{56.1}  & \colorbox{\paretocolor}{\textbf{\textit{12.04}}} \\
    &                                                           & 2     & 34.4      & 57.5  & 24.09 \\
  
    & AGCN$^\dagger$                                            & 1     & \colorbox{\paretocolor}{35.0}      & \colorbox{\paretocolor}{57.5}  & \colorbox{\paretocolor}{13.45} \\
    &                                                           & 2     & \colorbox{\paretocolor}{\textbf{\textit{36.9}}}      & \colorbox{\paretocolor}{59.6}  & \colorbox{\paretocolor}{26.91} \\
    & S-TR$^\dagger$                                            & 1     
                                                                        & \colorbox{\paretocolor}{32.0}      & \colorbox{\paretocolor}{54.9}  &  \colorbox{\paretocolor}{\textbf{11.62}} \\
    &                                                           & 2     & \colorbox{\paretocolor}{34.7}      & \colorbox{\paretocolor}{57.9}  & \colorbox{\paretocolor}{23.24} \\
  
    \midrule
    Frame 
    & \textit{Co}ST-GCN (ours)                                  & 1     & 31.8      & 54.6  & 0.16 \\
    &                                                           & 2     & 33.1      & 56.1  & 0.32 \\
    & \textit{Co}ST-GCN$^*$ (ours)                              & 1     & \colorbox{\paretocolor}{30.2}      & \colorbox{\paretocolor}{52.4}  &  \colorbox{\paretocolor}{\textbf{0.11}} \\
    &                                                           & 2     & \colorbox{\paretocolor}{32.2}      & \colorbox{\paretocolor}{54.5}  & \colorbox{\paretocolor}{0.22} \\
    & \textit{Co}AGCN (ours)                                    & 1     & \colorbox{\paretocolor}{33.0}      & \colorbox{\paretocolor}{55.5}  & \colorbox{\paretocolor}{0.18} \\
    &                                                           & 2     & \colorbox{\paretocolor}{\textbf{35.0}}      & \colorbox{\paretocolor}{\textbf{57.3}}  & \colorbox{\paretocolor}{0.36} \\
    & \textit{Co}AGCN$^*$ (ours)                                & 1     & 23.3      & 44.3  & \textbf{\textit{0.12}} \\
    &                                                           & 2     & 27.5      & 49.1  & 0.25 \\
    & \textit{Co}S-TR (ours)                                    & 1     & 29.7      & 52.6  & 0.16 \\
    &                                                           & 2     & \colorbox{\paretocolor}{\textbf{\textit{32.7}}}      & \colorbox{\paretocolor}{\textbf{\textit{55.6}}}  & \colorbox{\paretocolor}{0.31} \\
    & \textit{Co}S-TR$^*$ (ours)                                & 1     & 27.4      & 49.7  & \textbf{0.11} \\
    &                                                           & 2     & 29.9      & 52.7  & 0.22 \\
    \bottomrule
\end{tabular}
}
\end{center}
\vspace{-10pt}
\end{table}

\subsection{Comparison with Prior Works}

% Comparison with other frame-by-frame models.
Most current state-of-the-art methods for skeleton-based action recognition are not able to efficiently perform frame-by-frame predictions in the online setting, since they are constrained to operate on whole skeleton-sequences.
Some RNN-based methods, e.g. Deep-LSTM~\cite{Shahroudy_2016_NTURGBD} and VA-LSTM~\cite{zhang2017view}, can be used for redundancy-free frame-wise predictions, but their reported accuracy has been sub-par relative to newer methods that sprung from ST-GCN.
The recently proposed AGC-LSTM~\cite{si2019attention} does report results on-par with CNN-based methods, and might also be able to provide redundancy-free frame-wise results, but we cannot validate this due to the lack of publicly available source code and details in the published paper.

{While ShiftGCN and ShiftGCN++ offer impressively low FLOPs, it should be noted that the shift operation is not accounted for by the FLOPs metric. The FLOP count of shift-based methods are therefore low compared to other methods and may not reflect on-hardware performance. Due to the temporal shifts back and forth in time,  ShiftGCN and ShiftGCN++ cannot be easily transformed into Continual Inference Networks in their current form, though a \textit{Continual} Shift operation could be devised, where temporal shifts only occur backwards in time. Nevertheless, ShiftGCN++ offers a remarkable accuracy/FLOPs trade-off, which was attained using multiple complementary techniques, including architecture search and knowledge distillation. Instead of proposing one particular architecture, our proposed approach of translating ST-GCN-based networks into CINs is generic and can be applied alongside many other methods.}

Many works have shown that the inclusion of multiple modalities leads to increased accuracy~\cite{yan2018spatial, shi2019two, shi2019skeleton_directed, cheng2020skeleton, liu2020disentangling}. 
In our context, these modalities amount to \textit{joints}, which are the original coordinates of the body joints, and \textit{bones}, which are the differences between connected joints. Additional \textit{joint motion} and \textit{bone motion} modalities can be retrieved by computing the differences between adjacent frames in time for the joint and bone streams respectively.
Models are trained individually on each stream and combined by adding their softmax outputs prior to prediction.

We evaluate and compare our proposed continual models, which are \textit{Co}ST-GCN, \textit{Co}AGCN, \textit{Co}S-TR, with prior works on the NTU RGB+D 60, NTU RGB+D 120, and Kinetics Skeleton 400 datasets as presented in Tables \ref{tab:benchmark-ntu60}, \ref{tab:benchmark-ntu-120}, and \ref{tab:benchmark-kinetics}.

The \textit{Co}ST-GCN and \textit{Co}S-TR models transfer well across all datasets both with ($^*$) and without padding and stride modifications. For \textit{Co}AGCN, we find that the change to stride one deteriorates accuracy. We surmise that the attention matrix in \cref{eq:agc2} may need a larger receptive field (basing the attention on more nodes as in AGCN) to provide beneficial adaptations; a per-step change in attention might provide more noise than clarity in middle and late network layers.
As found in prior works, the multi-stream approach with ensemble predictions gives a meaningful boost in accuracy across all experiment. 

The Continual Skeleton models provide competitive accuracy at multiple orders of magnitude reduction of FLOPs per prediction in the online setting compared to the original non-continual models. While none of our results beat prior state-of-the-art accuracy in absolute terms, this was never the intent with the method. Rather, we have successfully shown that online inference can be greatly accelerated for models in the ST-GCN family with state-of-the-art accuracy/complexity trade-offs to follow. 
For instance, our one and two-stream \textit{Co}S-TR$^*$ achieve Pareto optimal\footnote{A solution is pareto optimal if there is no other solution where one metric is improved without causing deterioration in another.} results on all subsets of the NTU RGB+D 60 and NTU RGB+D 120 datasets meaning that no other model improves on either accuracy and FLOPs without reducing the other. Pareto optimal models have been highlighted in Tables \ref{tab:benchmark-ntu60}, \ref{tab:benchmark-ntu-120}, and \ref{tab:benchmark-kinetics} accordingly.
Our approach may be used similarly to accelerate other architectures for skeleton-based human action recognition with temporal convolutions.%, though a removal of temporal attention may pose a challenge, as we observed for \textit{Co}AGCN$^*$.

%@@@@@@@@@@@@@@@@@@@@@@@@@@@@@@@@@@@@@@@@@@@@@@@@@@@@@
%@@@@@@@@@@@@@@@@@@@@@@@@@@@@@@@@@@@@@@@@@@@@@@@@@@@@@

\section{Conclusion} \label{sec:conclusion}

In this paper, we proposed \textit{Continual} Spatio-Temporal Graph Convolutional Networks, an architectural enhancement for
networks operating on time-dependent graph structures, 
which augments prior methods with the ability to perform predictions frame-by-frame during online inference while attaining weight compatibility for batch inference. We re-implement and benchmark three prominent methods {for skeleton-based action recognition}, the ST-GCN, AGCN, and S-TR, as novel Continual Inference Networks, \textit{Co}ST-GCN, \textit{Co}AGCN, and \textit{Co}S-TR, and propose architectural modifications to maximize their frame-by-frame inference speed.
Through experiments on three widely used human skeleton datasets, NTU RGB+D 60, NTU RGB+D 120, and Kinetics Skeleton 400, we show up to 26$\times$ on-hardware throughput increases, 109$\times$ reduction in FLOPs per prediction, and 52\% reduction in maximum memory allocated during online inference with similar accuracy to those of the original networks.
{During offline inference of full spatio-temporal sequences, the models operate identically to prior works.}
Our proposed architectural modifications are generic in nature and can be used for a variety of problems involving time-dependent graph structures, like traffic control. 
Proposal of methods based on \textit{Continual} ST-GCNs which can address challenges posed by such problems can be an interesting future research direction.
{It is our hope, that this innovation will make online processing of time-varying graphs viable} on recourse-constrained devices and systems with real-time requirements.
\section*{Acknowledgment}
This work has received funding from the European Union’s Horizon 2020 research and innovation programme under grant agreement No 871449 (OpenDR). This publication reflects the authors’ views only. The European Commission is not responsible for any use that may be made of the information it contains.


%\section*{Acknowledgment}




\renewcommand*{\bibfont}{\small}
\bibliographystyle{IEEEtranN}
\bibliography{references.bib}


\begin{IEEEbiography}[{\includegraphics[width=1in,height=1.25in,clip,keepaspectratio]{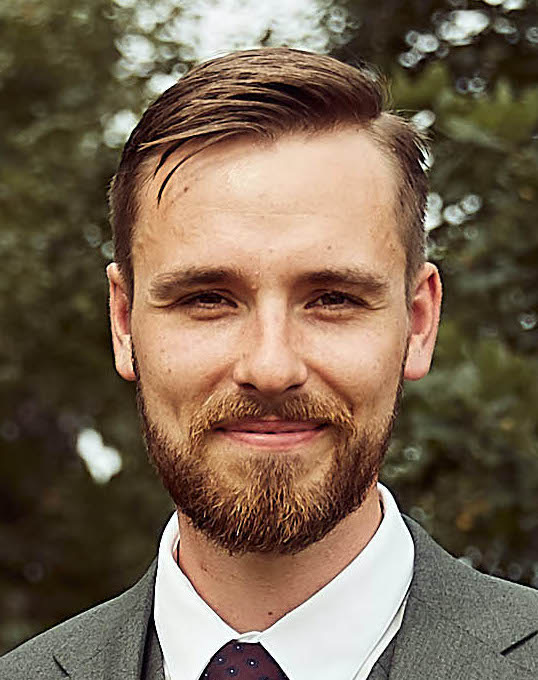}}]{Lukas Hedegaard}
is a PhD candidate at Aarhus University, Denmark. He received his M.Sc. degree in Computer Engineering in 2019 and B.Eng. degree in Electronics in 2017 at Aarhus University, specialising in signal processing and machine learning. With a common theme of efficient deep learning, his research endeavours span from online inference acceleration and human activity recognition to transfer learning and domain adaptation.
\end{IEEEbiography}

% \begin{IEEEbiographynophoto}{Name here}

\begin{IEEEbiography}[{\includegraphics[width=1in,height=1.25in,clip,keepaspectratio]{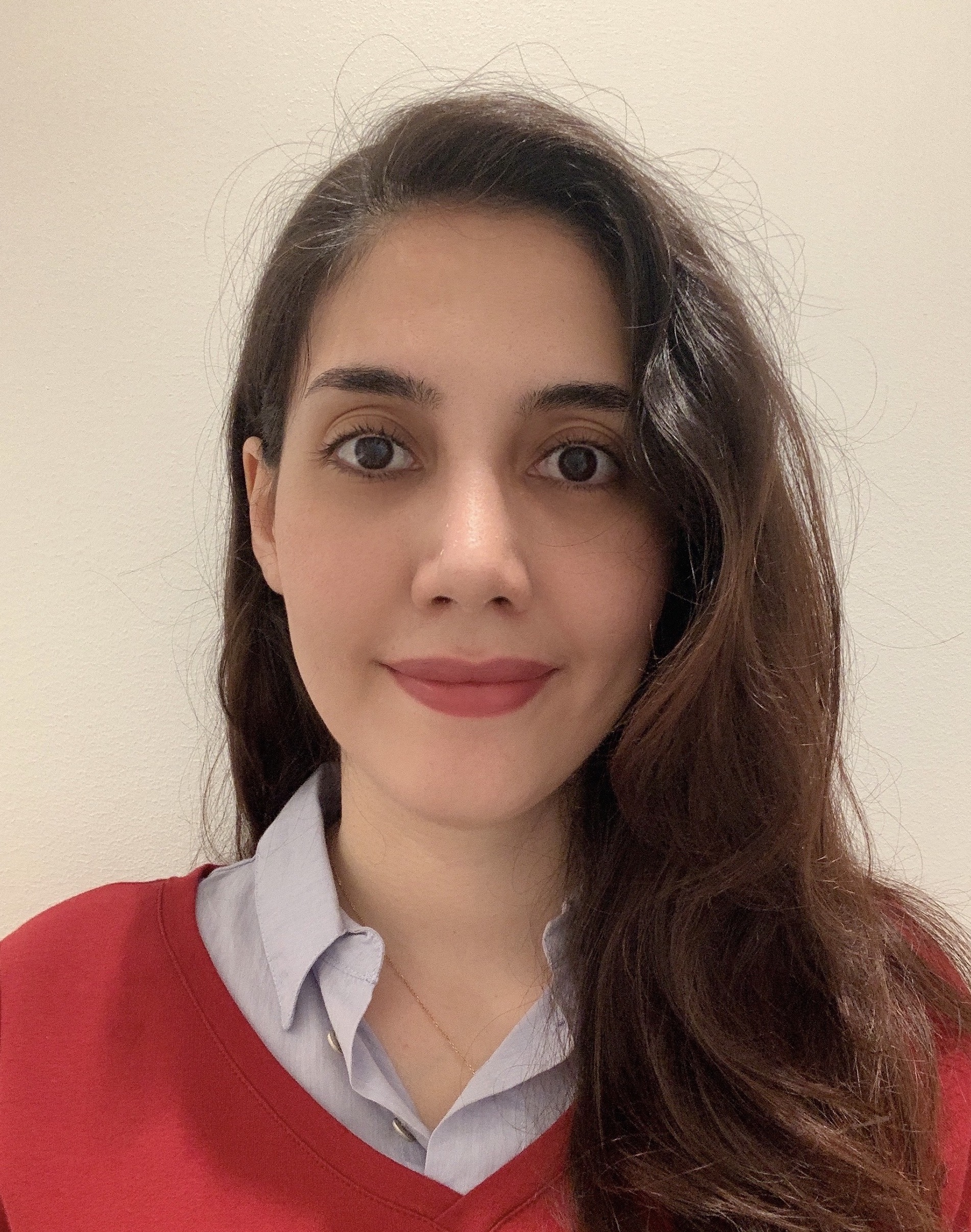}}]{Negar Heidari}
is a Postdoctoral researcher at Aarhus University, Denmark. She completed her PhD in Signal Processing and Machine Learning at the Department of Electrical and Computer Engineering, Aarhus University in 2022. Her current research interests include machine learning, deep learning and computer vision with a focus on computational efficiency. 

\end{IEEEbiography}

\begin{IEEEbiography}[{\includegraphics[width=1in,height=1.25in,clip,keepaspectratio]{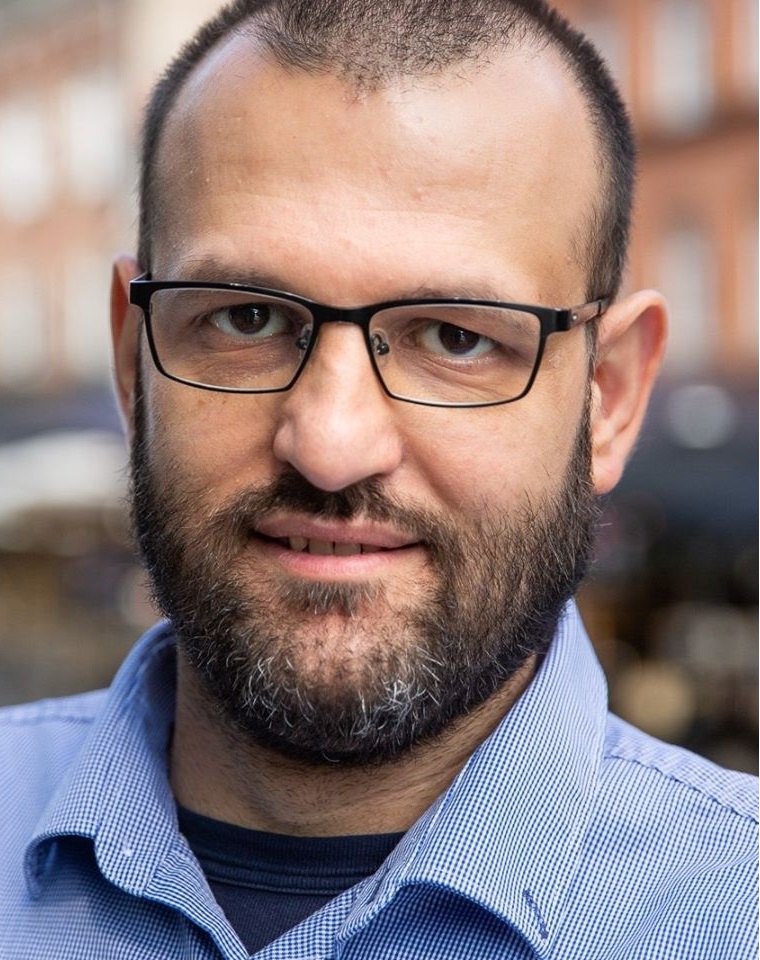}}]{Alexandros Iosifidis}
(SM'16) is a Professor at Aarhus University, Denmark. His research interests focus on neural networks and statistical machine learning finding applications in computer vision, financial modelling and graph analysis problems. He serves as Associate Editor in Chief for Neurocomputing (for Neural Networks research area), as an Area Editor for Signal Processing: Image Communication, and as an Associate Editor for IEEE Transactions on Neural Networks and Learning Systems. He was an Area Chair for IEEE ICIP 2018-2022 and EUSIPCO 2019,2021, and Publicity co-Chair of IEEE ICME 2021. He was the recipient of the EURASIP Early Career Award 2021 for contributions to statistical machine learning and artificial neural networks. \end{IEEEbiography}

\vfill


%\vfill


\end{document}